%% file: ijhcs-2017.tex
\documentclass[preprint]{elsarticle}

\usepackage{hyperref}

\journal{International Journal of Human-Computer Studies}

\usepackage{booktabs}
\usepackage{framed}
\usepackage{subfig}
\usepackage{amsmath}
\usepackage{rotating}

\usepackage{todonotes} 
\let\chapter\section 
\usepackage[ruled,noline,noend,linesnumbered]{algorithm2e}

\begin{document}

\begin{frontmatter}

\title{Explaining Reputation Assessments}

\author[ufrgs,tudortmund]{Ingrid Nunes\corref{cor1}} %
\ead{ingridnunes@inf.ufrgs.br} %
\cortext[cor1]{Corresponding author.} %
\author[warwick]{Phillip Taylor} %
\ead{Phillip.Taylor@warwick.ac.uk} %
\author[essex]{Lina Barakat} %
\ead{lina.barakat@essex.ac.uk} %
\author[warwick]{Nathan Griffiths} %
\ead{nathan.griffiths@warwick.ac.uk} %
\author[kings]{Simon Miles} %
\ead{simon.miles@kcl.ac.uk} %

\address[ufrgs]{Universidade Federal do Rio Grande do Sul (UFRGS), Porto Alegre, Brazil}
\address[tudortmund]{TU Dortmund, Dortmund, Germany}
\address[warwick]{University of Warwick, Coventry, United Kingdom}
\address[essex]{University of Essex, Colchester, United Kingdom}
\address[kings]{King's College London, London, United Kingdom}

\begin{abstract}
Reputation is crucial to enabling human or software agents to select among alternative providers. Although several effective reputation assessment methods exist, they typically distil reputation into a numerical representation, with no accompanying explanation of the rationale behind the assessment. Such explanations would allow users or clients to make a richer assessment of providers, and tailor selection according to their preferences and current context. In this paper, we propose an approach to explain the rationale behind assessments from quantitative reputation models, by generating arguments that are combined to form explanations. Our approach adapts, extends and combines existing approaches for explaining decisions made using multi-attribute decision models in the context of reputation. We present example argument templates, and describe how to select their parameters using explanation algorithms. Our proposal was evaluated by means of a user study, which followed an existing protocol. Our results give evidence that although explanations present a subset of the information of trust scores, they are sufficient to equally evaluate providers recommended based on their trust score. Moreover, when explanation arguments reveal implicit model information, they are less persuasive than scores. 
\end{abstract}


\end{frontmatter}

\input{sec1_introduction}
\input{sec2_relatedWork}

\input{sec3_mttm}
\input{sec4_explanation}
\input{sec5_evaluation}
\input{sec6_conclusion}

\section*{Acknowledgements}
This work was part funded by the UK Engineering and Physical Sciences Research Council as part of the Justified Assessments of Service Provider Reputation project, ref.\ EP/M012654/1 and EP/M012662/1. Ingrid Nunes thanks for research grants CNPq ref.\ 303232/2015-3, CAPES ref.\ 7619-15-4, and Alexander von Humboldt, ref.\ BRA 1184533 HFSTCAPES-P.

\section*{References}
\bibliography{jaspr,reputation}

\end{document}

%% file: sec1_introduction.tex

\section{Introduction}

In environments where many parties offer comparable services or products, customers need to be able to choose between the options available. Automated support for this has been studied extensively in the areas of recommender systems~\cite{Ricci:2010:RSH:1941884} and reputation assessment~\cite{Ramchurn:2004aa}. In particular, reputation assessment allows the calculation of reputation scores so that the past performance of service providers can be compared. These scores can then be used to determine which provider to select, as they characterise providers according to the factors of interest to the client. Various reputation models~\cite{Sabater:2004aa,Huynh:2006aa,Teacy:2005aa,Regan:2006aa,Teacy:2012aa} have been shown to be effective through empirical evaluation, but do not provide the transparency needed to understand why one provider has a better reputation than another. 
As the complexity of reputation models increases, this understanding is becoming harder to achieve.
Access to the reasons that underlie reputation assessment would allow users to judge whether the resulting reputation scores reflect their actual interests in the current context, and allow providers to identify the aspects they must improve. Explanations have been exploited to improve user system acceptance in expert systems and recommender systems~\cite{Tintarev:chapter2011:Explanation}, but have not been explored in the context where automated interactions occur, such as in multi-agent systems, or instantiated for reputation assessment methods.

Our goal is to improve, from the user perspective, the \emph{transparency} of reputation models, which are in general purely quantitative. Reputation scores are helpful to assess and rank providers but, with explanations of such scores, users would be able to evaluate whether they agree with them. As a consequence, users can make more \emph{effective} choices when taking reputation into account. We propose an approach to explain the rationale behind the scores generated by reputation assessment models. These are abstracted into a generic reputation model, which we refer to as the \emph{multi-term reputation model} (MTRM). This is not a new reputation model, but rather is a generalised model in which we can express existing reputation assessment methods, upon which explanations can be built. Our approach generates arguments about the reasons behind reputation scores by leveraging explanation approaches proposed in the context of multi-attribute utility theory~\cite{Labreuche:AI2011:Explanations,Nunes:ECAI2014:ExplanationTechnique}, and combines the arguments into explanations. Explanations are produced based on information that can be obtained from an instance of MTRM. Moreover, this generic reputation model can be customised, leading to an instantiation of a specific underlying existing reputation model, and model-specific arguments can then be generated. In order to illustrate this process, we show customisations for the FIRE~\cite{Huynh:2006aa} and TRAVOS~\cite{Teacy:2005aa} reputation models. 

Despite the fact that users have generally been taken out of the loop in evaluations of work on trust and reputation for multi-agent systems, a study involving real people is essential for validating our approach. Therefore, in order to evaluate our generated explanations, we conducted a user study, which provides evidence of their usefulness. The study involve 30 participants and followed the protocol proposed by Bilgic and Mooney~\cite{bilgic:iui05-wkshp}. As result, we observed that, in order to assess providers, our explanations is as efficient as having detailed information about trust scores of providers, that is, with less information (and possibly more confidentially) participants were able to assess providers. Furthermore, our explanation arguments are less persuasive than scores when they reveal implicit model information. In our study, arguments were presented to participants in a textual form, generated using example templates of how to transform our explanation arguments into a user-understandable form. This choice caused participants, however, to prefer trust scores, which were presented in a table, over textual explanations.

In summary, our key contribution is an approach to explain quantitative reputation models, focusing on FIRE and TRAVOS as illustrative reputation models. Specifically, we (i) propose a method to generate explanations of assessments from quantitative reputation models, (ii) show how to leverage existing approaches for explaining decisions made using multi-attribute decision models in the context of reputation, and (iii) evaluate such explanations through a user study.

We describe background research and related work in Section~\ref{sec:relatedWork}. The multi-term reputation model (MTRM) is introduced in Section~\ref{sec:MTRM}, followed by a description of our explanation approach in Section~\ref{sec:approach}. The user study performed to evaluate our approach is presented in Section~\ref{sec:evaluation}. Finally, we present our conclusions in Section~\ref{sec:conclusion}. 

%% file: sec2_relatedWork.tex

\section{Background and Related Work}\label{sec:relatedWork}

Two main research areas are associated with our work, namely, explanations for recommender and decision support systems, and trust and reputation assessment methods. There is much work that has been done in the former, but not addressing our particular context. We give an overview of explanation approaches and introduce those that are adopted in our work in Section~\ref{sec:RW:explanation}. Trust and reputation have also been widely investigated and, as a result, many reputation models have been proposed. Our approach aims to be generic, in the sense that it can be used with any reputation model. We instantiate it for illustration using two existing reputation models, FIRE~\cite{Huynh:2006aa} and TRAVOS~\cite{Teacy:2005aa,Teacy:2006aa}, as described in Section~\ref{sec:RW:reputation}.

\subsection{Explanation Generation} \label{sec:RW:explanation}

Over recent years, there has been an increasing interest in explanations for recommender and decision support systems~\cite{Tintarev:chapter2011:Explanation,Herlocker:CSCW2000:ExplainingRecommender,Carenini:AI2006:EvaluativeArguments}. Explanations in such systems have been investigated, as was the case with expert systems~\cite{Ye:MISQ1995:ExplanationImpact}, because explanations can promote many benefits, including increased user trust and more effective decisions~\cite{Tintarev:chapter2011:Explanation}, which are fundamental to user acceptance of these systems.

Different studies have been performed in the context of explanations. Many types of explanations given for recommender systems were compared in user studies~\cite{Herlocker:CSCW2000:ExplainingRecommender,Gedikli:2014:IEC:2580118.2580448}. Herlocker et al.~\cite{Herlocker:CSCW2000:ExplainingRecommender} concluded that showing rates from neighbours in the context of collaborative filtering (using histograms) contributes to the acceptance of the recommendation. However, Bilgic and Mooney~\cite{bilgic:iui05-wkshp} observed that this kind of explanation persuades users to accept recommendations rather than helping them to make better choices. Indeed, explanations can be given with different purposes~\cite{Tintarev:chapter2011:Explanation}. As Bilgic and Mooney argue, persuasion explanations cause users to overestimate the quality of an option and make inaccurate choices and, consequently, their confidence in the system rapidly deteriorates. Our interest is thus in \emph{effective} explanations \cite{Tintarev:chapter2011:Explanation}, which assist users to make better decisions by helping them to evaluate the quality of options according to their own preferences. There are some studies with people that give foundation to this kind of explanation \cite{Tintarev:RecSys2007:EffectiveExplanations,Nunes:2012:IEJ:2358968.2358987}, with the proposal of patterns and guidelines, which state that attributes presented in explanations must be tailored to the user, as has been confirmed by a previous user study \cite{Carenini:IJCAI2001:ExplanationStudy}.

There are three main approaches that propose algorithms that select attributes to be part of effective explanations~\cite{Klein:AA1994:ExplanationFW,Labreuche:AI2011:Explanations,Nunes:ECAI2014:ExplanationTechnique}. Such approaches use multi-attribute decision models as input, which makes them inadequate to be used as is with reputation models. However, they can be used in a complementary way in our work, by being adapted to be used in our context.

The oldest approach, proposed by Klein and Shortliffe~\cite{Klein:AA1994:ExplanationFW}, is empirically motivated but lacks proper evaluation, while Labreuche's approach \cite{Labreuche:AI2011:Explanations} addresses a limitation of this method---a formal justification of the selected arguments. Labreuche~\cite{Labreuche:AI2011:Explanations} proposed an approach for selecting and generating arguments for the family of multi-attribute decision models parameterised by weights assigned to the criteria, such as the expected utility model and the weighted majority model. The explanations generated are of four different types, generated using different kinds of argumentation reasoning, called \emph{anchors} (\emph{all}, \emph{not on average}, \emph{invert} and \emph{remaining case}). Anchors identify changes in a weight vector $v$ that yields an inversion of the prescription made by the decision model, leading to why one option is preferred to another. Two strategies for the modification of the weights are considered: (i) the replacement of $v$ by some reference weights $w^{\mathcal{F}}$, indicating that an option is preferred to another because it is better for the most important attributes, but not on average, and (ii) a permutation of the weights $v$ among the criteria (associated with a branch-and-bound algorithm), indicating that the preferred option is better for the most important attributes and worse for the least important attributes. A trivial anchor addresses the case of domination (the case where an option has at least one advantage with respect to another, and no disadvantage), and another last anchor covers the remaining cases.

An explanation generation technique was proposed by Nunes et al.~\cite{Nunes:ECAI2014:ExplanationTechnique}, which is founded on a study of how people justify choices~\cite{Nunes:2012:IEJ:2358968.2358987}. The technique is composed of a set of algorithms that select attributes to be used as part of explanations that follow different explanation patterns, such as \emph{critical attribute}, \emph{cut-off value}, \emph{decisive criteria} and \emph{trade-off resolution}. While Klein and Shortliffe's approach selects outlier attributes and Labreuche analyses weight changes, Nunes et al.\  consider a set of attributes as decisive when they are the minimum set of attributes (in the sense of $\subset$) needed to make an option worse than another. If this set consists of all cons of an option, then a second set of attributes is selected: the minimum set of attributes that are pros that must not be taken into account to enable the existence of a decisive criteria.

We have used adapted parts of these two introduced approaches~\cite{Labreuche:AI2011:Explanations,Nunes:ECAI2014:ExplanationTechnique} in the work described in this paper, and further details of these parts are provided when we describe our explanation approach.

Argumentation frameworks have also been adopted for the purpose of empowering quantitative decision tools with inference mechanisms and respective explanation capabilities---e.g.\ argumentation-enriched recommender systems have been proposed for recommending music~\cite{Briguez12}, movies~\cite{RecioGarcia13, BRIGUEZ14}, web content~\cite{Chesnevar09}, and learning objects~\cite{Rodriguez16}. In many such approaches, Defeasible Logic Programming (DeLP)~\cite{Garcia04} is employed either instead, or on top of an existing quantitative technique in order to provide a qualitative perspective, where conclusions/suggestions are reasoned in terms of arguments for and against them. In particular, DeLP models (potentially inconsistent and contradictory) knowledge about the domain, in terms of facts and a set of strict and defeasible inference rules. An argument for a particular conclusion/suggestion is then derived by applying backward chaining on these facts and rules. Arguments can be attacked by other arguments (e.g.\ those proposing opposite conclusions), and the attacks among arguments can be resolved via associating arguments with probabilities/preferences.

The knowledge (facts and rules) upon which the reasoning of such argumentation frameworks is based is typically pre-determined, and is derived directly from user preference declarations, and added on top of the (sub-)results of the quantitative measure. Our explanation approach focuses on providing a finer-grained analysis of the reasoning behind the quantitative measure (rather than substituting it or building on top of it), and can be seen as a dynamic generator of knowledge to then be used by such argumentation frameworks.

\input{sec2_relatedWork_intro_and_fire}

\input{sec2_relatedWork_travos}

%% file: sec2_relatedWork_intro_and_fire.tex

\subsection{Reputation Models} \label{sec:RW:reputation}

Trust and reputation enable agents to minimise the inherent uncertainty when self-interested individuals or organisations interact~\cite{Sabater:2005aa}. Trust can be viewed as an assessment of the likelihood that an individual or organisation will fulfil its commitments~\cite{Gambetta:1988aa}. Reputation complements trust, and can be seen as a public perception of trustworthiness~\cite{Josang:2007aa}. Several computational models of trust and reputation exist, which can be broadly categorised into those that are based on credentials and those based on experience and observation of past behaviour---see~\cite{Sabater:2005aa,Josang:2007aa,Ramchurn:2004aa,Pinyol:2013aa} for comprehensive reviews. Credential-based approaches use policies to express when, for what, and how to determine trust based on  certificates, keys, or digital signatures, etc. Although such methods are effective for managing access rights and permissions, they do not support more general reasoning about interactions, and therefore in this paper we focus instead on experience based approaches.

Several experience based approaches
use a combination of direct and indirect experience to derive a numerical or probabilistic assessment of reputation~\cite{Wang:2007aa}. ReGreT~\cite{Sabater-Mir:2001aa,Sabater:2004aa} assesses reputation on three aspects: (i) an individual dimension from direct experience, (ii) a social dimension using knowledge of others' experiences and the social structure, and (iii) an ontological dimension that accounts for the different aspects that inform reputation (e.g.\ delivery, price, and quality). FIRE~\cite{Huynh:2006aa} builds on ReGreT through the addition of role-based trust, and certified reputation based on third-party references~\cite{Huynh:2006aa}. TRAVOS~\cite{Teacy:2005aa} takes a probabilistic approach to assessing trust, estimating the expected value of success of future interactions using a beta probability distribution.

The use of a binary variable (success or failure) to model outcomes is a limitation of TRAVOS and alternative approaches have been proposed. For example, BLADE~\cite{Regan:2006aa} models agents and advisor evaluation functions as dynamic random variables using Dirichlet distributions, enabling progressive learning of probabilistic models through Bayesian techniques. 
To cope with noisy advisors, HABIT~\cite{Teacy:2012aa} creates a Bayesian network to support reasoning about reputation. However, HABIT assumes that the distribution of an agent's behaviour is static, an assumption not made by other approaches.
Other reputation systems apply machine learning in assessing reputation, typically in assessing stereotypical reputation~\cite{Burnett:2013aa,Sensoy:2016aa}.

Although these methods rely on different aggregations/distributions, they have been used for the same purpose of estimating the reputation of agents with which an agent wants to interact, relying on evaluations made based on previous interactions (either by direct experience or with peers) over time. In this paper, we adopt FIRE and TRAVOS as examples to illustrate our approach, and describe their operation in more detail below. We focus on FIRE and TRAVOS due to their simplicity and low computational overheads, compared to approaches such as BLADE and HABIT, because the focus of this paper is on explanation generation providing a rationale for reputation assessment, rather than on any particular reputation assessment method itself. We selected two methods to demonstrate the generality of our approach and the value of customisations made to particular methods. 

\subsection{The FIRE Reputation Model}\label{sec:FIRE}

FIRE combines four types of reputation and trust: interaction trust from direct experience ($I$), witness reputation from third party reports ($W$), role-based trust ($R$), and certified reputation based on third-party references ($Cr$)~\cite{Huynh:2006aa}.
Reputation is assessed in FIRE from {\em rating} tuples, $(a, b, t, i, v)$, where $a$ and $b$ are agents that participated in interaction $i$ such that $a$ gave $b$ a rating value of $v \in [-1,+1]$ for the term $t$ (e.g. reliability, quality, timeliness). 
A rating of $+1$ is absolutely positive, $-1$ is absolutely negative, and $0$ is neutral.  
In FIRE, each agent has a history of size $H$ and stores the last $H$ ratings it has given in its local database.
FIRE gives more weight to recent interactions using a {\em rating weight function}, $\omega_K$, for each trust or reputation component $K \in \{I, W, R, Cr\}$.

The component trust or reputation $a$ has in $b$ for term $t$ is the weighted mean of ratings,
\begin{equation}
\label{weightedMean}
\mathcal{T}_K (a,b,t) = \frac{\sum_{r_i \in \mathcal{R}_K (a, b, t)} \omega_K (r_i) \cdot v_i}{\sum_{r_i \in \mathcal{R}_K (a, b, t)} \omega_K (r_i)}
\end{equation}
where $\mathcal{R}_K (a, b, t)$ is the set of ratings stored by $a$ regarding $b$ for component $K$ with respect to term $t$, and $v_i$ is the value of rating $r_i$.
Interaction trust, $\mathcal{T}_I (a,b,t)$ is calculated from the interaction records that the assessing agent $a$ has in their database, $\mathcal{R}_I(a, b, t)$. 
Specifically, the ratings of records matching $(a, b, t, \_, \_)$ are aggregated using Equation~\ref{weightedMean}, where $b$ is the agent being assessed, $t$ is the term of interest, and ``$\_$'' matches any value, and:
\begin{equation}
\label{ratingWeight_I}
\omega_I(r_i) = e^{-\frac{\Delta \tau(r_i)}{\lambda}}
\end{equation}
Here, $\omega_I(r_i)$ is the weight for rating $r_i$ and $\Delta \tau(r_i)$ is the time since rating $r_i$ was recorded.

Witness and certified reputation are similarly calculated, using this aggregation over different sets of interaction ratings.
For witness reputation the assessing agent, $a$, uses a acquaintances to provide their ratings of $b$ for term $t$, i.e.\ ratings of the form $(\_, b, t, \_, \_)$.
If the acquaintance has no relevant experience, they will pass on the request to their own acquaintances.
To assess certified reputation, the assessed agent, $b$, provides a set of ratings that they have previously been given by other agents.
The weighting used in calculating witness and certified reputation is $\omega_W(r_i) = \omega_Cr(r_i) = \omega_I(r_i)$.

Role-based trust uses ratings assigned to rules describing agent relationships, e.g.\ if they are part of the same organisation, or there is a provider consumer relationship. 
Rules have the form $(role_a, role_b, t, e, v)$, representing if two agents $a$ and $b$ take roles $role_a$ and $role_b$, then $b$ is expected with a likelihood of $e \in [0,1]$ to have performance of $v$ for term $t$ in an interaction with $a$.
To calculate role-based trust, rules in the assessing agent's database that match $\mathcal{R}_R (a, b, t)$ are aggregated using Equation~\ref{weightedMean}, with $\omega_R(r_i) = e_i$.

The composite term trust, $\mathcal{T} (a,b,t)$, in an agent with respect to a given term $t$ is calculated as a weighted mean of the component sources:
\begin{equation}
\label{termTrust}
\mathcal{T} (a,b,t) = \frac{\sum_{K \in \{I, W, R, Cr\}} \omega_K \cdot \mathcal{T}_K (a, b, t)}{\sum_{K \in \{I, W, R, Cr\}} \omega_K}
\end{equation}
where $\omega_I$, $\omega_W$, $\omega_R$ and $\omega_{Cr}$ are parameters that determine the importance of each component, $\omega_K = \omega_K \cdot \rho_K (a, b, t)$, and the reliability of the reputation value for component $K$ is $ \rho_K (a, b, t)$.
The reliability of a reputation value is determined by a combination of the rating reliability and rating deviation reliability (details of the calculations can be found in~\cite{Huynh:2006aa}).

%% file: sec2_relatedWork_travos.tex
\subsection{The TRAVOS Reputation System}

TRAVOS is based on the Beta Reputation System~\cite{Whitby:2004aa} and extends it to ignore reputation assessments from unreliable witnesses \cite{Teacy:2005aa,Josang:2002aa,Teacy:2006aa}.
TRAVOS uses interaction trust and witness reputation, computed using rating tuples similar to those used in FIRE.
Whereas in FIRE the rating value is a real number, ratings in TRAVOS are binary, $v \in \{0,1\}$, where $0$ is a negative rating and $1$ is positive.
The component trust value agent $a$ has in agent $b$ with respect to term $t$, is the expected value of a beta probability density function,
\begin{equation}
\mathcal{T}_{K}(a,b,t) = \frac{\alpha_{K}(a,b,t)}{\alpha_{K}(a,b,t)+\beta_{K}(a,b,t)},
\end{equation}
where $\alpha_{K}(a,b,t)$ is $1$ plus the number of relevant positive ratings and $\beta_{K}(a,b,t)$ is $1$ plus the number of relevant negative ratings,
\begin{equation}
\begin{aligned}
\alpha_{K}(a,b,t) &= 1 + |\{r_i \in \mathcal{R}_{K}(a, b, t) | v_{i} = 1\}|,
\\
\beta_{K}(a,b,t) &= 1 + |\{r_i \in \mathcal{R}_{K}(a, b, t) | v_{i} = 0\}|.
\end{aligned}
\end{equation}

The beta probability density function can also be used to compute a confidence in the trust value, defined by the proportion of the distribution that lies in a range centred around the expected value,
\begin{equation}
\mathcal{\rho}_{K}(a,b,t) =
\frac{
	\int_{\mathcal{T}_{K}(a,b,t)-\epsilon}^{\mathcal{T}_{K}(a,b,t)+\epsilon}
	X^{\alpha_{K}(a,b,t)-1}(1-X)^{\beta_{K}(a,b,t)-1}dX
}
{\int_0^1 U^{\alpha_{K}(a,b,t)-1}(1-U)^{\beta_{K}(a,b,t)-1}dU},
\end{equation}
where $\epsilon$ is a user defined parameter to define the range considered.

As with FIRE, an assessing agent computes interaction trust from the set of ratings, $\mathcal{R}_{I}(a, b, t)$, in its database that match $(a, b, t, \_, \_)$.
The interaction trust is then $\mathcal{T}_{I}(a,b,t)$, which has an associated confidence, $\mathcal{\rho}_{I}(a,b,t)$.
If $\mathcal{\rho}_{I}(a,b,t)$ is below a threshold set by the user, witnesses are asked for ratings of agent $b$ for term $t$, which are used to compute the witness reputation.

Witnesses, $w \in W$, provide opinions in the form of the number of positive, $\alpha_{W}(w,b,t)$ and the number of negative ratings, $\beta_{W}(w,b,t)$, that they have given $b$.
Before the overall reputation is calculated, the witness opinions are discounted based on their perceived accuracy to limit their effect on the composite reputation score.
TRAVOS stores previous ratings provided by witnesses in \emph{observation} tuples, $(a,w,b,t,i,o,v)$, where, $w$ is a witness that provided evaluator $a$ with a set of ratings about provider $b$, which formed a beta probability density distribution whose expected value determined the raw opinion value of $o$.
After processing this witness opinion and selecting $b$ to interact with, $a$ gave $b$ a rating value of $v$ in interaction $i$.

On receipt of a new opinion from a witness, $w$, an evaluator, $a$, queries their observation database for records where the opinion, $o$, provided by $w$ for term $t$ was similar.
Two opinions are said to be similar if their expected values are close (i.e.\ they both lie in the same discrete interval).
The coherence of the opinion provided, $o$, and the rating for the subsequent interaction, $v$, then determines the reliability of the new opinion provided by the witness.
Given this reliability, the opinion is discounted and combined along with the interaction trust by summing the $\alpha$ and $\beta$ parameters,
\begin{equation}\label{eq:travos_comb}
\begin{aligned}
\alpha(a,b,t) &= \alpha_{I}(a,b,t) + \sum_{w \in W}\bar{\alpha}_{W}(w,b,t)
\\
\beta(a,b,t) &= \beta_{I}(a,b,t) + \sum_{w \in W}\bar{\beta}_{W}(w,b,t),
\end{aligned}
\end{equation}
where $\bar{\alpha}_{W}(w,b,t)$ and $\bar{\beta}_{W}(w,b,t)$ are the discounted opinion parameters provided by witness $w$ regarding agent $b$ for term $t$.
The composite term trust in agent $b$ for term $t$ is then,
\begin{equation}
\mathcal{T}(a,b,t) = \frac{\alpha(a,b,t)}{\alpha(a,b,t)+\beta(a,b,t)}.
\end{equation}
For full details on the calculation behind discounting see~\cite{Teacy:2006aa}.

%% file: sec3_mttm.tex
\section{Multi-Term Reputation Model}\label{sec:MTRM}

In the previous section, we gave an overview of two different reputation models, namely FIRE and TRAVOS. In order to provide a model-independent explanation approach, we must first specify a common model specification that generalises different reputation models. This generalised model, which we refer to as \emph{multi-term reputation model} (MTRM), can be specialised by the addition of the specific components of a particular reputation model. Note this MTRM is not a new reputation model, but a model that captures concepts present in \emph{any} reputation model. Therefore, explanations provided based on this model are applicable to any reputation model. Concepts that are usual, e.g.\ recency, but not used in all reputation models can be added in MTRM extensions. We next introduce the MTRM concepts.

All reputation models consider a way for an agent to assess how an interaction with another agent occurred. In FIRE, for example, agents associate a rating with those they interact with in $[-1,+1]$, while in TRAVOS agents only record success or failure, i.e.\ ratings are in $\{0,1\}$. These ratings are then communicated to others who require additional information to inform their decisions. In our model, we consider that an agent is associated with a set of trust ratings
\begin{equation}
\label{eq:trust_rating}
r_i = \langle a, b, t, K, v \rangle
\end{equation}
where $a$ is a source agent, $b$ is target agent, $t$ is a term, $K$ is a reputation type, and $v$ is a rating value. A particular reputation model may add additional parameters, e.g.\ interaction as in FIRE. Trust ratings are associated with reputation types, or components, according to the component of the model that generates them. Each model incorporates a particular set of reputation types, $K_{Set}$. TRAVOS only includes interaction ($I$) and witness ($W$) reputation types, while FIRE supplements those with role-based ($R$) and certified ($Cr$) reputation. The set of ratings associated with a particular reputation type is $\mathcal{R}_K (a, b, t)$.

These ratings are used to calculate a trust value $\mathcal{T}_K(a,b,t)$, which combines trust ratings in a single real value. In case of FIRE, as introduced in Section~\ref{sec:FIRE}, the trust value is a weighted mean of ratings, considering a recency function $\omega_\lambda(r_i)$, while TRAVOS uses a probabilistic model. If a trust value is associated with a reputation type $K$, it means that it is derived from ratings only associated with $K$.

Trust values associated with different reputation types must be combined to form a single value. In MTRM, as its name indicates, we consider that agents can assess others with respect to different terms $t \in T$, such as cost, quality and timeliness. The component trust values can be combined to form the term trust $\mathcal{T}(a,b,t)$.
We do not assume that the term trust is calculated using a specific method such as a weighted mean or sum, but rather we assume that the term trust can be decomposed into weights $\omega_K$ and trust values $\mathcal{T}_K(a,b,t)$, associated with different reputation types. This is straightforward in FIRE, given that FIRE calculates trust as a weighted mean of weighted means. However, TRAVOS does not calculate a composite trust value from interaction and witness trusts in this way, instead combining ratings from witnesses, after adjustment for reliability and relevance, to act as parameters of a beta probability distribution whose expected value determines the composite trust value. Consequently, we use the TRAVOS model to compute the term trust from ratings, and then decompose this term trust into two trust values, one associated with direct interaction trust, and another with witness trust.

TRAVOS computes an overall trust value, which in our case is the term trust, by combining the direct interaction trust and witness opinions, after adjusting them for perceived accuracy.
The combination proceeds by summing the $\alpha$ and $\beta$ parameters of the beta probability density functions, as in Equation~\ref{eq:travos_comb}.
In FIRE, the trust component weights are determined by user preferences, while in TRAVOS, we define them as the proportion of the final beta probability density function that the component parameters account for.
For instance, the interaction trust weight is,
\begin{equation}
\omega_I(a,b,t) = \frac{\alpha_I(a,b,t)+\beta_I(a,b,t)}{\alpha_I(a,b,t) + \beta_I(a,b,t) + \sum_{w \in W}\bar{\alpha}_W(w,b,t) + \bar{\beta}_W(w,b,t)}
\end{equation}
and witness reputation weight is $\omega_W(a,b,t) = 1 - \omega_I(a,b,t)$.

Finally, existing reputation models either do not consider terms (e.g.\ TRAVOS) or often do not specify how to combine values for different terms into a single trust score (as is the case with FIRE). Therefore, inspired by multi-attribute utility theory~\cite{Keeney:book1976:MAUT}, we consider weights that establish a trade-off relationship among terms, and view term trust as a utility value. The overall trust score is then a weighted mean of term trusts, where the weights are agents' preferences for terms.
\begin{equation}
\label{overallTrust}
\mathcal{T} (a,b) = \frac{\sum_{t \in T} \omega_t \cdot \mathcal{T} (a, b, t)} {\sum_{t \in T} \omega_t}
\end{equation}
where the parameters $\omega_t$ correspond to $a$'s preferences regarding the relative importance of terms, and $T$ is the set of all terms.




Note that in order for reputation models to be abstracted to our MTRM, they should either use a weighted sum approach, like FIRE, or be decomposable into such an approach, like TRAVOS. 

As result, our MTRM is able to capture data such as that presented in Table~\ref{table:runningExample}. In this table, we show a set of illustrative trust values from the perspective of an agent $A$ with respect to four other agents ($B$, $C$, $D$ and $E$), considering three different terms---Quality (Q), Timeliness (T) and Cost (Ct). For example, $\mathcal{T}_I(A,D,T) = 0.95$. These trust values are combinations of ratings by, for example, a recency function. Similarly, there are trust values that come from witnesses, which are shown in the columns labelled with ``Witness Trust Values'' in Table~\ref{table:runningExample}, for instance $\mathcal{T}_W(A,C,Q) = 0.40$.

\begin{table}
  \centering
	\scriptsize
  \begin{tabular}{ l|r r r|r r r|r r r|r}
    \toprule
     & \multicolumn{3}{c|}{\textbf{Interaction}} & \multicolumn{3}{c|}{\textbf{Witness}}  & \multicolumn{3}{c|}{}  &  \\ 
     & \multicolumn{3}{c|}{\textbf{Trust Values}} & \multicolumn{3}{c|}{\textbf{Trust Values}}  & \multicolumn{3}{c|}{\textbf{Term Trusts}}  & \textbf{Trust Score} \\  \cline{2-10}
		 & \textbf{Q} & \textbf{T} & \textbf{Ct} & \textbf{Q} & \textbf{T} & \textbf{Ct} & \textbf{Q} & \textbf{T} & \textbf{Ct} & \\
		\midrule
		\textbf{B}      & 0.75 & 0.55 & 0.40 & 0.95 & 0.70 & 0.30 & 0.80 & 0.59 & 0.38 & 0.64 \\ 
		\textbf{C}      & 0.10 & 0.20 & 0.15 & 0.40 & 0.15 & 0.15 & 0.18 & 0.19 & 0.15 & 0.17 \\ 
		\textbf{D}      & 0.50 & 0.95 & 0.10 & 0.60 & 0.80 & 0.10 & 0.53 & 0.91 & 0.10 & 0.58 \\ 
		\textbf{E}      & 0.10 & 0.20 & 0.40 & 0.90 & 1.00 & 0.95 & 0.30 & 0.40 & 0.54 & 0.38 \\ 
		\midrule
		\textbf{Weight} & 0.45 & 0.35 & 0.20 &      &      &      &      &      &       & \\ 
		\bottomrule
  \end{tabular}
  \caption{Running Example: Agents and Scores.} %
  \label{table:runningExample}
\end{table}

The term trust, in this case, combines interaction and witness trust values. Assume that agent $A$ uses the following weights: (i) interaction weight: $\omega_I = 0.75$; and (ii) witness weight: $\omega_W = 0.25$. As result, for example, we have the quality trust with respect to $B$ would be $\mathcal{T} (A, B, Q) = 0.75 \times 0.75 + 0.25 \times 0.95 = 0.80$. 

Similarly, term trusts are combined using weights, which are shown in the last row of Table~\ref{table:runningExample}, for terms resulting in the overall trust score, shown in the last column in Table~\ref{table:runningExample}---for instance,  $\mathcal{T} (A, C) = 0.17$. Based on these calculations, it can be seen that the agent with the best trust score is agent $B$. Although there is a mathematical explanation that leads to this, it is hard to extract intuitive arguments that justify why $B$ is the most trustworthy agent for agent $A$. This is done by our explanation approach, which is presented in the following section.

%% file: sec4_explanation.tex

\section{Explaining Reputation Assessments}\label{sec:approach}

Now that we have a common reputation model, we can specify a method for producing explanations. An explanation justifies why a particular agent (e.g.\ a service provider) has a better reputation, i.e.\ the overall trust score, than another from the perspective of a given agent (e.g.\ a client). Our explanations are produced by generating a set of arguments, which give the key aspects that distinguish the two agents being compared, being all arguments needed to understand which agent is better. Arguments are instantiated with parameters selected using specified algorithms. We first present arguments that can be part of an explanation, and then show how to use these arguments to produce an explanation.

Our method not only produces arguments for our common trust model, MTRM, but also considers the specific details of different reputation models. Therefore we have generic arguments, generated based on MTRM, which are supplemented with model-specific arguments. We show as examples of the latter specific arguments for both FIRE and TRAVOS, which are used as illustrative reputation assessment models in this paper.

\subsection{Explanation Arguments}\label{sec:arguments}

We first look at the possible classes of reasons why a provider may have a better reputation than another. Such classes are associated with the different components that are part of MTRM. Each class has a corresponding argument type that can be used as part of an explanation. The generation of arguments here is similar to the identification of decisive criteria to explain choices made using multi-attribute decision models. We select, adapt and combine the algorithms of Labreuche~\cite{Labreuche:AI2011:Explanations} and Nunes et al.~\cite{Nunes:ECAI2014:ExplanationTechnique} to produce our arguments. As described earlier, an agent's overall trust score is a weighted mean of term trust values, and each of these can be decomposed into trust values for different reputation types. Correspondingly, our argument types are split into three groups, namely decisive terms, decisive reputation types, and reputation model-specific arguments, as described below. For simplicity, but without loss of generality, we assume that ratings are in $[0,1]$, given that the approaches we leverage use this range. FIRE and TRAVOS ratings can be easily mapped to this range.

\subsubsection{Argument: Decisive Terms}

The reputation of a provider for a client is a balance among trust values for terms, corresponding to aspects of an interaction or service such as quality or timeliness. Some terms may be irrelevant with respect to why one provider is more trusted than another, either because they have low weight for the client or because the differences between term trust values for providers are small. To explain why provider $b$ has a better overall trust score than provider $b'$ for an agent $a$, we must identify the \emph{decisive} terms $\mathcal{D}(a,b,b') = \langle P, C \rangle$ that lead to this conclusion, where $P$ and $C$ are sets of terms that are the decisive \emph{pros} and \emph{cons} of $b$ with respect to $b'$, respectively. For example, if $P = \{ quality, cost \}$ and $C = \{timeliness \}$, we can derive an argument of the form ``\emph{$b$ is more trusted than $b'$ because it has higher trust for quality and cost, even though $b'$ has higher trust for timeliness}''.

A trivial case is that of \emph{domination}, when $b$ has advantages compared to $b'$ with respect to some terms and no disadvantages with respect to the remaining terms. According to Labreuche, important terms are those that have weights higher than the reference weight, which is defined as the weight that makes all terms equality important (used in the \emph{not on average} anchor, $\Psi_{NOA}$). That is, if there are $n$ terms, the reference weight is $\omega^\mathcal{A} = 1/n$. We need to adapt this to take into account the trust values for terms. Considering the difference between term trust for a term $t$ for providers $b$ and $b'$, $\Delta_t = | \mathcal{T}(a,b,t) - \mathcal{T}(a,b',t) |$, we can say that the reference value difference is $\Delta^\mathcal{A} = \frac{\sum_{t \in T} \Delta_t}{|T|}$, where $T$ is the set of terms. Thus, $\Delta^\mathcal{A}$ is the average of the differences between trust values for all terms. Given the reference weight and reference value difference, the reference weighted value difference is $\omega^\mathcal{A} \cdot \Delta^\mathcal{A}$. Decisive terms in the case of domination are consequently those whose weighted value difference is higher than the reference weighted value difference, i.e.\
\begin{equation}
\mathcal{D}_{Dom}(a,b,b') = \langle \{ t \in T | \omega_t\cdot\Delta_t >  \omega^\mathcal{A}\cdot\Delta^\mathcal{A} \} , \emptyset \rangle
\end{equation}
Informally, decisive pros are terms that have: (i) above average weight and value, (ii) very high weight, or (iii) very high value. In this context ``\emph{very high}''
 means that even though $\Delta_t < \Delta^\mathcal{A}$, $\omega_t$ is high enough to cause $\omega_t\cdot\Delta_t >  \omega^\mathcal{A}\cdot\Delta^\mathcal{A}$, and the same reasoning is applied to $\Delta_t$.  As provider $b$ dominates $b'$, there are no cons in this case.

In order to illustrate the domination case, we use the example introduced in the previous section, considering the values presented in Table~\ref{table:runningExample}. By analysing the term trusts of agents $B$ and $C$, it is possible to see that $B$ dominates $C$, because $B$ has higher trust values for all terms. In order to identify the decisive terms, we first calculate the reference value difference, which is
\[
\Delta^\mathcal{A} = \frac{| 0.80 - 0.18 | + | 0.59 - 0.19 | + | 0.38 - 0.15 |}{3} = \frac{0.63 + 0.40 + 0.23}{3} = 0.42
\]
As $\omega^\mathcal{A} = 0.33$, $\omega^\mathcal{A}\cdot\Delta^\mathcal{A} = 0.14$. Calculating the weighted differences for quality, timeliness and costs, we obtain 0.28, 0.14 and 0.05, respectively. As only the first two are above the reference weighted value difference\footnote{The reference weighted value is, more precisely, 0.139.}, they are the decisive terms. An explanation argument, in this case, would be as follows.

\begin{framed} \footnotesize
\textbf{Example 1:} $B$ has a better reputation than $C$, because it is better in all aspects that you consider in your preferences, mainly with respect to timeliness, and quality.
\end{framed}

When dominance is not the case, we could apply either Labreuche's anchors~\cite{Labreuche:AI2011:Explanations} or the patterns of Nunes et al.~\cite{Nunes:ECAI2014:ExplanationTechnique} to select decisive criteria. As the number of terms $|T|$ may be high and Labreuche's approach may have performance issues~\cite{Nunes:ECAI2014:ExplanationTechnique}, we use the latter, which is briefly explained as follows.
We first define $T_+ = \{ t \in T | \mathcal{T}(a,b,t) > \mathcal{T}(a,b',t) \}$ and $T_- = \{ t \in T | \mathcal{T}(a,b,t) < \mathcal{T}(a,b',t) \}$, which are the sets of all pros and cons of $b$ with respect to $b'$, respectively. Using these patterns, the decisive criteria is
$\mathcal{D}_{DC}(a,b,b') = \langle T_{+}^{*}, T_{-}^{*} \rangle$,
such that
$T_{+}^{*} \subseteq T_{+}$, $T_{-}^{*} \subseteq T_{-}$, and
\begin{equation}
\sum_{t \in T_{+}^{*}} \omega_t\cdot\Delta_t > \sum_{t \in T_{-}/T_{-}^{*}} \omega_t\cdot\Delta_t
\end{equation}
$T_{+}^{*}$ and $T_{-}^{*}$ are both minimal in the sense of $\subseteq$. When $T_{-}^{*} = \emptyset$, it is a decisive criteria pattern, otherwise it is a trade-off resolution pattern.

In order to better understand the selection of decisive terms when there is no dominance, we use our running example. Consider agents $B$ and $D$. According to the trust value, the former has two pros, namely quality (weighted difference is $0.12$) and cost (weighted difference is $0.06$), while the latter has only timeliness (weighted difference is $0.11$) as pros. In order to justify why $D$ is less trustworthy than $B$, considering only quality would be enough, because its weighted difference is already higher than the weighted difference of timeliness (its con). Therefore, quality is $B$'s decisive criteria with respect to $D$. This is illustrated in the argument below.

\begin{framed}\footnotesize
\textbf{Example 2:} $B$ has a better reputation than $D$, mainly due to quality.
\end{framed}

\subsubsection{Argument: Decisive Reputation Types}

The key argument produced to explain why provider $b$ is more trusted than provider $b'$ is the set of terms that are the decisive pros of $b$ with respect to $b'$, and occasionally the decisive cons of $b'$. Term trusts are derived from ratings of different kinds of sources, referred to as reputation types, $K$, being a composition of trust values considering different sources. Therefore, we can again leverage algorithms used for multi-attribute decision models, to refine the explanation.

When $b$ dominates $b'$ for a term $t$, i.e.\ there exists $K$ in the set of reputation types such that $\mathcal{T}_K(a,b,t) > \mathcal{T}_K(a,b',t)$ and there is no $K'$ such that $\mathcal{T}_{K'}(a,b,t) < \mathcal{T}_{K'}(a,b',t)$, then stating that $t$ is a decisive term is sufficient, and no additional argument is needed. In other cases, it is relevant to add new arguments to the explanation. For example, assume that $b$ has a higher trust score than $b'$ considering a component $I$ (for interaction trust), $b'$ has a higher trust score than $b$ considering $W$ (for witness trust), and $\omega_I \gg \omega_W$ ($I$ is more important than $W$). In this case, it is helpful to state the argument  ``\emph{even though $b'$ has higher ratings from third party reports, $b$ has higher ratings from direct experience, which is more important}.''

Our pairwise analysis of weights and values is done with Labreuche's \emph{invert} anchor, $\Psi_{IVT}$. Although this anchor had performance issues in a previously performed experiment with human participants~\cite{Nunes:ECAI2014:ExplanationTechnique}, this occurred where there was a high number of attributes, which in our case corresponds to reputation types. We assume there is a small number of reputation types (e.g.\ there are four in FIRE and two in TRAVOS) and so performance is not an issue here. The argument given for explaining trust values considering reputation types is a permutation $\pi(a,b,b',t) = \{ (K,K') \in S^{2} \}$, where $S \subseteq K_{Set}$, such that $\mathcal{T}(a,b,t) <_{\pi(a,b,b',t)} \mathcal{T}(a,b',t)$. The operator $<_{\pi(a,b,b',t)}$ compares two term trusts applying the permutation $\pi(a,b,b',t)$ to reputation type weights. Consequently, $\pi(a,b,b',t)$ gives a set of pairwise changes in weights, which causes the term trust of $b'$ to be higher than that of $b$. Labreuche provides a branch-and-bound algorithm for the determination of this kind of explanation~\cite{Labreuche:AI2011:Explanations}, which for brevity is not reproduced here. Given that there are limited possible permutations in our case, algorithmic efficiency is not critical.

Considering our running example, we have a case of decisive reputation types considering agents $B$ and $E$ with respect to the timeliness term. The trust value of agent $B$ is better considering interaction ratings ($0.55 > 0.20$), while the trust value of agent $E$ is better considering witnesses ratings ($1.00 > 0.70$). If the weights given to the interaction and witnesses ratings were inverted, $E$ would have a higher term trust than $B$---timeliness trust would be 0.66 for $B$ and $0.80$ for $E$, instead of $0.59$ and $0.40$, respectively. We present below a textual argument that gives this explanation.

\begin{framed}\footnotesize
\textbf{Example 3:} Considering timeliness, even though $E$ has higher reputation with respect to witness reputation, which is less important, $B$ has has higher reputation with respect to own interaction, which is more important.
\end{framed}

\subsubsection{Reputation Model-specific Arguments}

The way that trust and reputation values are derived from ratings is different for each reputation model. As a consequence, it is possible to provide further arguments other than our generic arguments if we take model particularities into account. In this case, model-specific arguments can be generated and used to supplement the generic arguments. In addition, arguments can be added not only to explain trust scores, but to give further details about trust values and term trusts. Here, to illustrate these possibilities, we present two model-specific arguments: a FIRE-specific argument associated with trust values and a TRAVOS-specific argument to further explain term trusts.

\paragraph{FIRE-specific Argument: Recency}

The trust value for a particular reputation type in FIRE is calculated through a weighted mean of available ratings $v_i$.
Weights can be used to assign more importance to
particular ratings, specifically more recent ratings have a higher weight. The ratings are thus scaled using a rating recency factor $\lambda$, as introduced before. The recency factor may play a key role both in the overall trust score and in the trust value for particular $t$ and $K$. The overall trust score of a provider uses $\omega_\lambda(r_i)$ to combine available ratings $\mathcal{R}_K (a, b, t)$, associated with a particular $a$, $b$ and $t$. In this case, we can also consider a reference rating weight function $\omega_{\lambda_K}^\mathcal{A}$, which is the average weight, i.e.\
\begin{equation}
\omega_{\lambda_K}^\mathcal{A} = \frac{1}{|\mathcal{R}_K (a, b, t)|}
\end{equation}

Given this reference function, two situations might occur. First, the order derived from the overall trust score of providers $b$ and $b'$, calculated taking into account recency, conflicts with the order derived from the overall trust score calculated using $\omega_{\lambda_K}^\mathcal{A}$. That is, we have $\mathcal{T}(a,b) > \mathcal{T}(a,b')$ and $\mathcal{T}^{\mathcal{A}}(a,b) < \mathcal{T}^{\mathcal{A}}(a,b')$, where $\mathcal{T}^{\mathcal{A}}(a,b)$ is the overall trust calculated using $\omega_{\lambda_K}^\mathcal{A}$. Second, even though this situation may not occur, there may still be cases where $\mathcal{T}_K(a,b,t) > \mathcal{T}_K(a,b',t)$ and $\mathcal{T}^{\mathcal{A}}_K(a,b,t) < \mathcal{T}^{\mathcal{A}}_K(a,b',t)$, for a particular $K$ and $t$.
In the first scenario, we add an argument $\mathcal{F}(a,b,b')$ to the explanation explaining that ``\emph{although on average $b'$ has higher ratings than $b$, recently $b$ has been receiving higher ratings than $b'$, which are more valuable}''. In the second case, we must add a finer-grained argument $\mathcal{F}(a,b,b',t,K)$, for specific $K$ and $t$: ``\emph{although on average $b'$ has higher ratings for t than $b$, considering K, recently $b$ has been receiving higher ratings than $b'$, which are more valuable}''.

\paragraph{TRAVOS-specific Argument: Low Confidence}

FIRE uses weights of reputation types to express their importance for a particular assessor agent, and they remain fixed unless an assessor explicitly changes them. Therefore, a set of interaction and witness ratings does not influence the weights of reputation types to calculate a trust score. TRAVOS, on the other hand, evaluates how useful interaction ratings are, before taking witness ratings into account. If an assessor does not have enough confidence into its own ratings, i.e.\ the confidence is below a given threshold, then witness ratings are used, otherwise it will rely on its own ratings.

Therefore, it is important to know whether the trust score is based solely on interaction ratings or on both interaction and witness ratings. If $\mathcal{\rho}_{I}(a,b,t)$ (interaction confidence) is below a threshold set by the assessor, for any of the providers being assessed, it means that witness ratings are being taken into account to consider $b$ better than $b'$, i.e.\ $\mathcal{T}_W(a,b,t) > \mathcal{T}_W(a,b',t)$.
When this is the case, we add an argument $\mathcal{C}(a,b,b',t)$ to the explanation, which can be written in natural language in the following form: ``\emph{although you have had limited previous interactions with either $b$ or $b'$  with respect to $t$, the former is considered better than the latter by witnesses}''.




\subsection{Explanation Generation}\label{sec:explanation}

Above, we introduced the different arguments that can be used to form an explanation to justify why a provider $b$ has a higher trust score than a provider $b'$. In this section, we show how to generate such an explanation.
We first identify our coarse-grained argument to justify trust scores. This argument is composed of decisive terms, which has the form $\mathcal{D}(a,b,b')$ and gives the decisive pros and cons justifying the overall trust scores. When $b$ dominates $b'$, i.e.\ exists $t \in T$ such that $\mathcal{T}(a,b,t) > \mathcal{T}(a,b',t)$ and there is no $t' \in T$ such that $\mathcal{T}(a,b,t') < \mathcal{T}(a,b',t')$, the decisive criteria are given by $\mathcal{D}_{Dom}(a,b,b')$, otherwise they are given by $\mathcal{D}_{DC}(a,b,b')$.

Once we know the decisive criteria that justify trust scores, we can provide fine-grained arguments that provide further understanding, considering decisive terms $t \in P$. First, we search for those that have a trust score associated with decisive reputation types. This is given by $\pi(a,b,b',t)$, which is a permutation of weights given for the different reputation types, indicating that the weights involved in that permutation are decisive, because if they were assigned in a different way, we would have $\mathcal{T}(a,b,t) < \mathcal{T}(a,b',t)$. Second, we add model-specific arguments. For example, in the case of FIRE, the arguments $\mathcal{F}(a,b,b')$ and $\mathcal{F}(a,b,b',t,K)$ are added when the selected recency weight function is the cause for making the trust value of $b$ higher than that of $b'$, i.e.\ if equal weights were given to all ratings, this would not have been the case. While in the case of TRAVOS, the argument $\mathcal{C}(a,b,b',t)$ is added when interaction ratings are limited, and thus the opinions of witnesses are taken into account.

\begin{algorithm}[t]
    \scriptsize
    \SetKwFunction{Dominates}{dominates}
    \SetKwFunction{AddSpecificArguments}{addSpecificArguments}
    \SetKwFunction{AddSpecificTermTrustArguments}{addSpecificTermTrustArguments}
    \SetKwFunction{AddSpecificTrustValueArguments}{addSpecificTrustValueArguments}

    \KwIn{$a$: an agent; $b, b'$: service providers}
    \KwOut{$\phi$: explanation with a set of arguments}
    \BlankLine
    \If{\Dominates(b,b')}{
    	$\phi \leftarrow \{ \mathcal{D}_{Dom}(a,b,b') \}$\;
    }
    \Else{
    	$\phi \leftarrow \{ \mathcal{D}_{DC}(a,b,b') \}$\;
    }
    \AddSpecificArguments($\phi$)\;
    \ForEach{$t \in P$}{
    	\If{$\exists \pi(a,b,b',t)$ such that $\mathcal{T}(a,b,t) <_{\pi} \mathcal{T}(a,b',t)$}{
			$\phi \leftarrow \phi \cup \{ \pi(a,b,b',t) \}$;\
		}
    	\AddSpecificTermTrustArguments($\phi, t,\mathcal{T}(a,b,t),\mathcal{T}(a,b,t')$)\;
    	\ForEach{$K \in K_{Set}$}{
    		\AddSpecificTrustValueArguments($\phi, t,K,\mathcal{T}_K(a,b,t),\mathcal{T}_K(a,b,t')$)\;
		}
	}
    \Return{$\phi$}\;
\caption{Expl($a,b,b'$)}
\label{alg:explanation}
\end{algorithm}

This method is presented in Algorithm~\ref{alg:explanation}, which generates an explanation $Expl(a,b,b')$ to justify why provider $b$ has a higher trust score than provider $b'$, for agent $a$. An explanation is thus a set of arguments of the types introduced above. Note that in Algorithm~\ref{alg:explanation}, fine-grained arguments are generated only for terms that are decisive pros. However, arguments may be also generated for decisive cons, if one wants to provide  further details about the trust score. No fine-grained arguments are generated for the remaining terms, since they are not decisive. In addition, Algorithm~\ref{alg:explanation} calls functions that add additional arguments to the explanations. These functions must be specified for specific trust models. For example, in the case of FIRE we can add recency arguments to explain the trust score as a whole and particular trust values, as shown in Algorithms~\ref{alg:fireRecency} and~\ref{alg:fireRecencyTK}. Similarly, for TRAVOS we can add arguments to explain term trust, as shown in Algorithm~\ref{alg:IvtTravos}.

\begin{algorithm}[!t]
    \scriptsize

    \KwIn{$\phi$: explanation}
    \KwOut{$\phi$: explanation with added arguments}
    \BlankLine

    \If{$\mathcal{T}(a,b) > \mathcal{T}(a,b')$ and $\mathcal{T}^\mathcal{A}(a,b) < \mathcal{T}^\mathcal{A}(a,b')$}{
		$\phi \leftarrow \phi \cup \{ \mathcal{F}(a,b,b') \}$;\
	}

    \Return{$\phi$}\;
\caption{FIRE: addSpecificArguments}
\label{alg:fireRecency}
\end{algorithm}

\begin{algorithm}[!t]
    \scriptsize

    \KwIn{$\phi$: explanation; $t$: term; $K$: reputation type $\mathcal{T}_K(a,b,t)$, $\mathcal{T}_K(a,b',t)$: trust values}
    \KwOut{$\phi$: explanation with added arguments}
    \BlankLine

    \If{$\mathcal{T}_K(a,b,t) > \mathcal{T}_K(a,b',t)$ and $\mathcal{T}^{\mathcal{A}}_K(a,b,t) < \mathcal{T}^{\mathcal{A}}_K(a,b',t)$}{
		$\phi \leftarrow \phi \cup \{ \mathcal{F}(a,b,b',t,K) \}$;\
	}
    \Return{$\phi$}\;
\caption{FIRE: addSpecificTrustValueArguments}
\label{alg:fireRecencyTK}
\end{algorithm}

\begin{algorithm}[!t]
    \scriptsize

    \KwIn{$\phi$: explanation; $t$: term; $\mathcal{\rho}_{I}(a,b,t)$, $\mathcal{\rho}_{I}(a,b',t)$: confidence; $\mathcal{T}_W(a,b,t)$, $\mathcal{T}_W(a,b',t)$: trust values}
    \KwOut{$\phi$: explanation with added arguments}
    \BlankLine

    \If{($\mathcal{\rho}_{I}(a,b,t) < \epsilon$ or $\mathcal{\rho}_{I}(a,b',t) < \epsilon$) and $\mathcal{T}_W(a,b,t) > \mathcal{T}_W(a,b',t)$.}{
		$\phi \leftarrow \phi \cup \{ \mathcal{C}(a,b,b',t) \}$;\
	}
    \Return{$\phi$}\;
\caption{TRAVOS: addSpecificTermTrustArguments}
\label{alg:IvtTravos}
\end{algorithm}



Finally, we now show how an explanation $Expl(a,b,b')$, which is a set of arguments, can be translated to human-readable form. For illustration, we adopt a textual form. Parts shown in brackets are optional, and thus may not appear in all explanations. Note that two of the optional arguments are FIRE-specific and one is TRAVOS-specific. In addition, optional arguments may be added more than once, depending on the number of arguments that are part of the explanation.
\begin{framed} \footnotesize
\emph{\underline{~~Provider $b$~~} has a better reputation than \underline{~~Provider $b'$~~} mainly due to \underline{~~list of pros in $P$~~} [, even though \underline{~~Provider $b'$~~} provides better \underline{~~list of cons in $C$~~}]$_{C \neq \emptyset}$.}

\emph{[In addition, \underline{~~Provider $b'$~~} has, on average, higher ratings than \underline{~~Provider $b$~~}, but \underline{~~Provider $b$~~} has been recently receiving higher ratings than \underline{~~Provider $b'$~~}, which are more valuable.]$_{\mathcal{F}(a,b,b')}$}

\emph{[Considering \underline{~~Term $t$~~}, even though \underline{~~Provider $b'$~~} has a higher trust value considering \underline{~~Reputation Type $K$~~}, which is less important, \underline{~~Provider $b$~~} has a higher trust value considering \underline{~~Reputation Type $K'$~~}, which is more important.]$_{\forall (K,K') \in \pi(a,b,b',t)}$}

\emph{[Moreover, although you have had limited previous interactions with either \underline{~~Provider $b$~~} or \underline{~~Provider $b'$~~} with respect to \underline{~~Term $t$~~}, the former is considered better than the latter by witnesses.]$_{\forall \mathcal{C}(a,b,b',t)}$}


\emph{[Moreover, \underline{~~Provider $b'$~~} has, on average, higher ratings for \underline{~~Term $t$~~} than $b$, considering \underline{~~Reputation Type $K$~~}, but \underline{~~Provider $b$~~} has been recently receiving higher ratings than \underline{~~Provider $b'$~~}, which are more valuable.]$_{\forall \mathcal{F}(a,b,b',t,K)}$}
\end{framed}
%

%% file: sec5_evaluation.tex

\section{User Study}\label{sec:evaluation}

In this section, we therefore present a user study conducted to evaluate our proposed explanation approach.


\subsection{Goal and Research Questions}\label{sec:userstudy-goal-RQs}

Reputation assessment models are often used in multiagent systems to allow autonomous agents (which can be humans) to identify in which agents they can trust to interact with. Our explanations can be used as a means for agents to exchange information regarding the reputation of other agents, without the need for exposing the reputation model details or detailed scores. However, as our explanations reveal less information than components of trust scores, we must evaluate if they are helpful for agents or users to better choose another agent (which can be, e.g.\, a service provider) to interact with. More specifically, we aim to answer the following research questions.
\begin{enumerate}
\item Are our explanations more effective in helping users to understand reputation-based recommendations than quantitative scores alone?
\item How do users perceive the usefulness of  our explanations?
\end{enumerate}
In order to answer these questions, we present our explanations to users using our example explanation templates. Our hypothesis is that users are better able to understand the rationale behind recommendations when they receive explanations instead of only quantitative information (i.e.\ reputation scores). Our first research question is aligned with this hypothesis. However, given that the effectiveness of such explanations may be different to how users perceive their usefulness, the second research question aims to explore this relationship.


\subsection{Procedure}\label{sec:procedure}

Our user study followed an adaptation of the protocol previously adopted to conduct user studies that involve the evaluation of explanations in recommender systems~\cite{bilgic:iui05-wkshp,Gedikli:2014:IEC:2580118.2580448}. The steps of this protocol are the following~\cite{bilgic:iui05-wkshp}: (1) get sample ratings from the user; (2) compute a recommendation $r$; (3) for each explanation system, present $r$ to the user with $e$'s explanation and ask the user to rate $r$; and (4) ask the user to try $r$ and then rate it again. In the remainder of this section we present the steps we followed to conduct the user study.

\paragraph{Construction of Provider Model} Our study involves participants rating and receiving recommendations of service providers based on reputation models. In order to have a set of providers to be part of the study, we create a set of simulated providers. Providers are described with a model that specifies the probabilities of transaction outcomes, e.g.\ considering a provider of delivery services, an outcome is the number of days taken to deliver a package. Outcomes are associated with terms, e.g.\ the outcome of delivering a package is associated with the term timeliness.

\paragraph{Participant Data and Preference Elicitation} Participants initiate the study by providing data about themselves and preferences for different terms. Additionally, they provide preferences for reputation types, required by the FIRE model.

\begin{figure}[t]
	\centering
	\subfloat[Sample Ratings: Generated Outcome.]{\includegraphics[width=\linewidth]{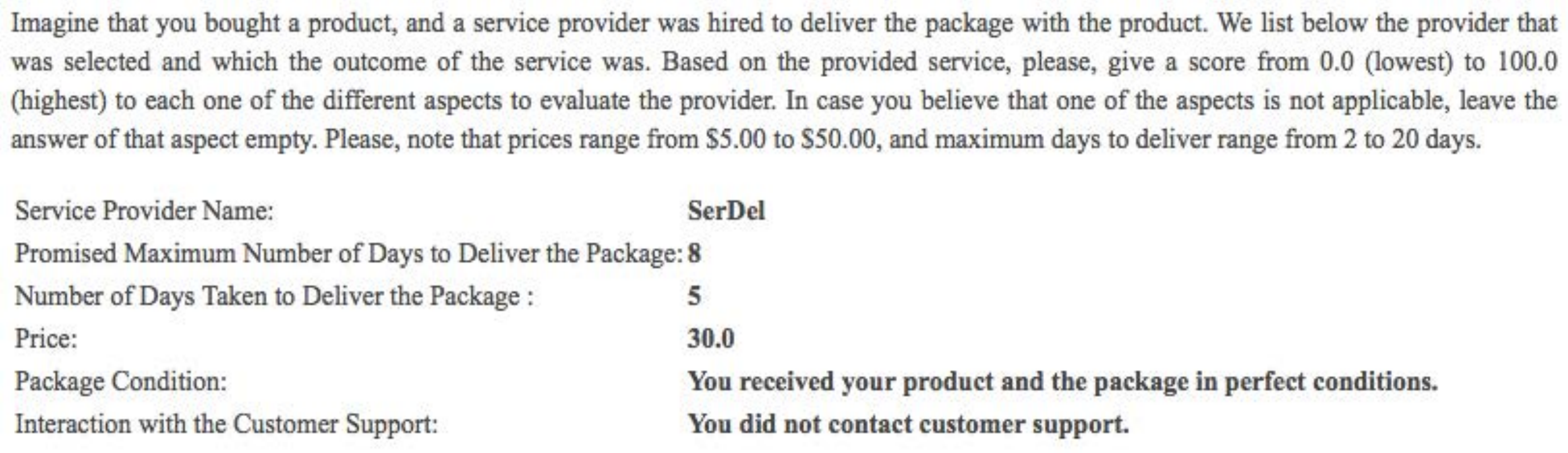}		
		\label{fig:screenshotSampleRatings}}\\
	\subfloat[Full Provider Information.]{\includegraphics[width=\linewidth]{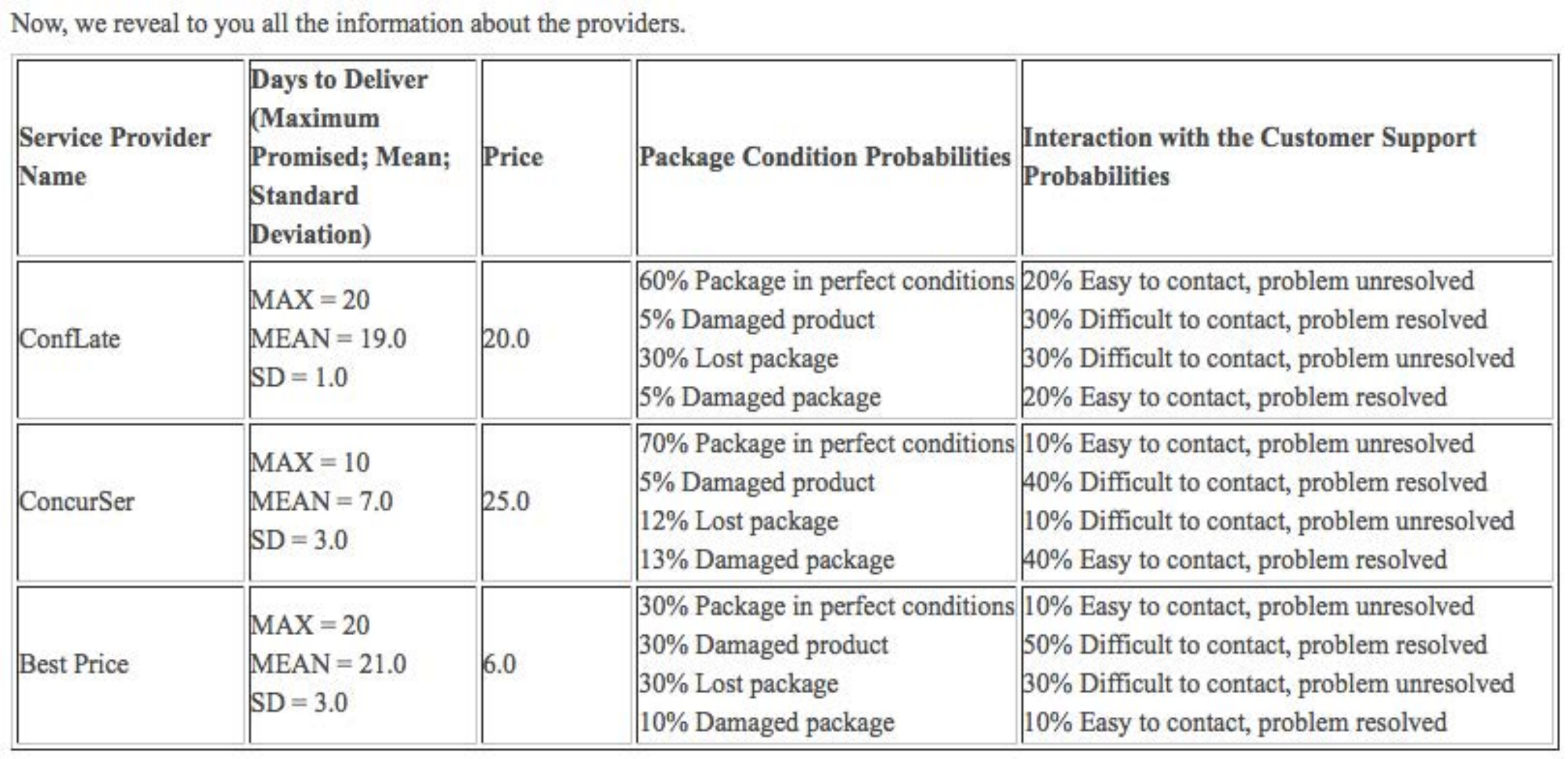}
		\label{fig:screenshotExplFullInfo}}
	\caption{Screenshots of the Web Application (1).}
	\label{fig:screenshots1}
\end{figure}

\paragraph{Collection of Sample Ratings} From each participant, we collect 15 sample ratings in the following way: (i) randomly select a provider, (ii) simulate an interaction by generating outcomes based on the provider model, and (iii) present the result of the interaction to the participant and ask them to rate the provider with respect to each term. We present an example of an interaction outcome in Figure~\ref{fig:screenshotSampleRatings}. Note that providers may be selected more than once, and likely have different outcomes in each interaction. Each set of ratings is associated with a round, which is interpreted as a timestamp for FIRE and a round for TRAVOS. These sample ratings are used to build both the FIRE and TRAVOS models for each participant. Participants provide ratings with a value between 0 and 1 (or not applicable). For FIRE, this value is used as is, and for TRAVOS we used a threshold of 0.5 to distinguish between successful and unsuccessful interactions. Moreover, TRAVOS requires a confidence threshold, which was set to 0.2. We selected a low threshold given that participants have few repeated experiences with the same provider, causing confidence to be usually low. In this way, we balance situations where witness opinions are used or not.

\begin{figure}[t]
	\centering
	\subfloat[Explanation Arguments.]{\includegraphics[width=\linewidth]{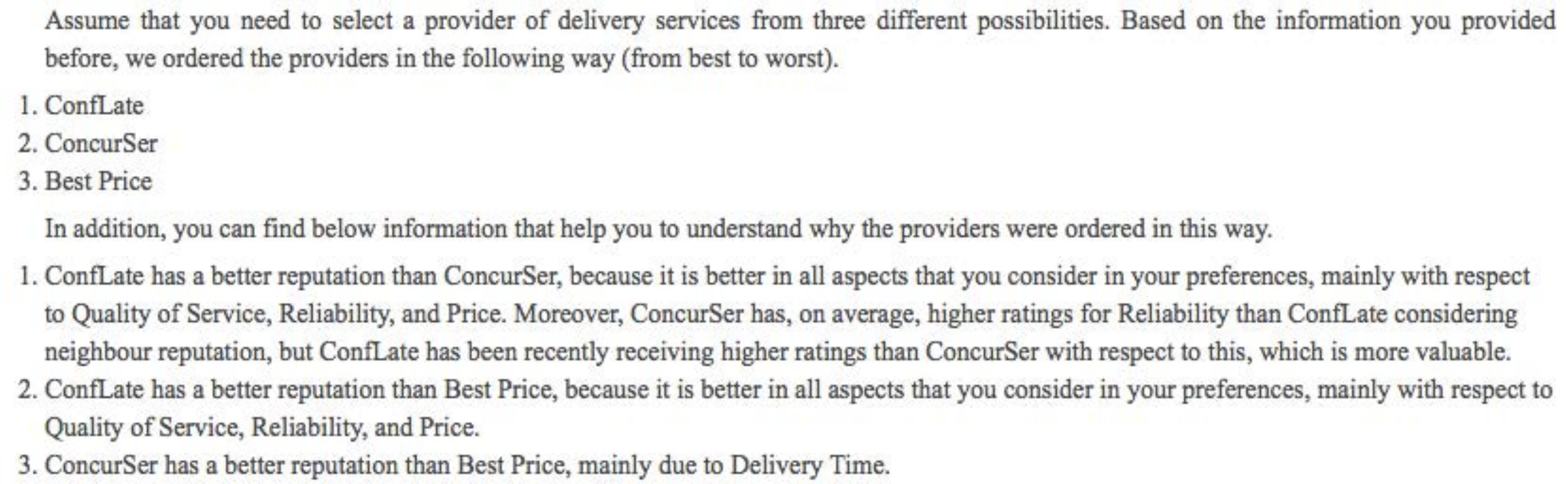}
		\label{fig:screenshotExplArguments}}\\
	\subfloat[Explanation Scores.]{\includegraphics[width=\linewidth]{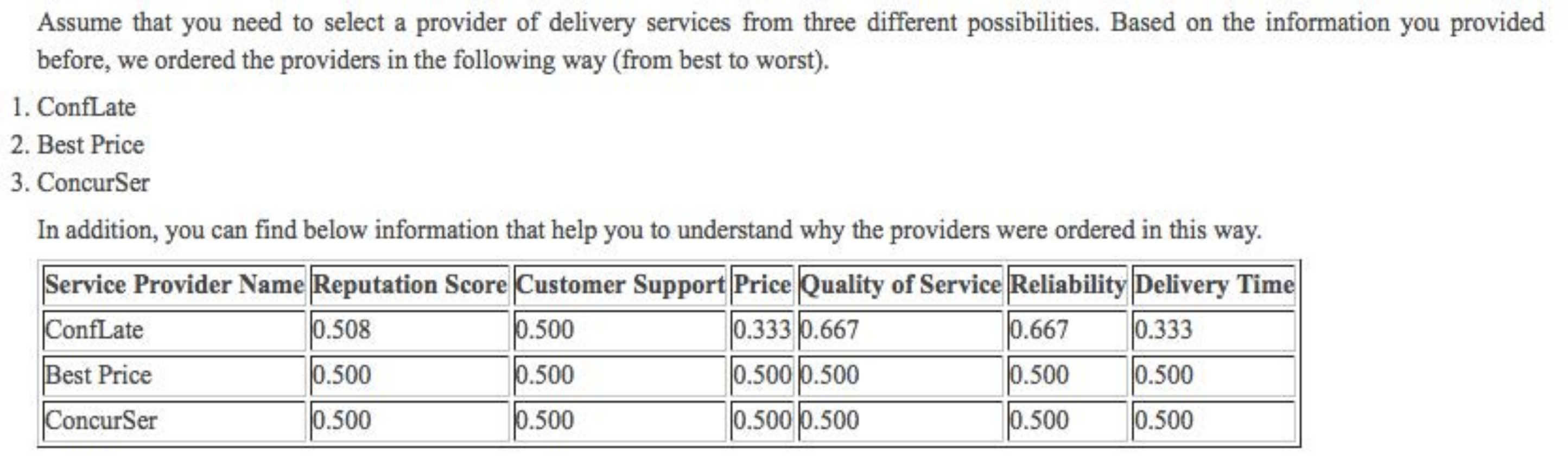}
		\label{fig:screenshotExplScores}}\\
	\caption{Screenshots of the Web Application (2).}
	\label{fig:screenshots2}
\end{figure}

\paragraph{Explanation Evaluation} We randomly select three providers from the set of providers and rank them using their computed reputation scores (step 2 of the protocol), which are based on the reputation model, ratings (from the participant and peers) and preferences. We randomly select the model to be used and which explanatory information is provided to users: (i) FIRE with scores alone, (ii) FIRE with explanation arguments alone, (iii) TRAVOS with scores alone, or (iv) TRAVOS with explanation arguments alone. Examples of explanation arguments and scores are shown in Figures~\ref{fig:screenshotExplArguments} and~\ref{fig:screenshotExplScores}, respectively. Note that participants are not aware that there are two underlying reputation models driving the recommendations. Then, we show to participants the provider ranking, together with the selected explanatory information (step 3 of the protocol), and ask them to answer in a 7-point Likert scale whether they agree with the statement: \emph{Considering the information provided above, I would order the presented providers in the same way that they were ordered, according to my preferences.} Next, we show participants the same ranking together with the full provider model (i.e.\ the probabilities of the outcomes), such as presented in Figure~\ref{fig:screenshotExplFullInfo}, so that they know all possible details about this provider, and ask them again the same question (step 4 of the protocol). Based on these answers we measure how the scores given for the first question (scores or explanation arguments) differ from the scores given for the provider model. With full information of providers' probabilities, participants know exactly what to expect by interacting with providers; however, this complete information is usually unknown. Therefore, the participant score with respect to full information is used as a baseline: the closer the participant score for explanation arguments or reputation scores, the better. This is therefore the metric we collect to evaluate the effectiveness of explanatory information, in the form of absolute difference between the two answers, referred to as \emph{score difference}. This step is repeated 10 times for each participant.

\paragraph{Perceived Effectiveness Questionnaire} To collect information regarding the \emph{perceived} value of the provided explanations, we ask participants to evaluate (in a 7-point Likert scale) the two forms of describing providers (with textual explanations and with reputation scores) with respect to (i) transparency: \emph{I understand why the providers were ranked in the presented way through the explanations} and  (ii) trust: \emph{I feel that these explanations are trustworthy}. In addition, we also ask an open-ended question to participants, in which participants have to explain their preference for scores or explanation arguments.

\subsection{Target Domain and Application Support}\label{sec:domainApp}

To execute the procedure described above, we implemented a web application to support the study, from which screenshots are presented in Figure~\ref{fig:screenshots1} and~\ref{fig:screenshots2}. We selected \emph{delivery services} as the domain, given that it is suitable for our scenario, because: (i) people in general have used this kind of service at least once and, if not, they are aware of how it works and its possible outcomes, and (ii) participants do not need to \emph{concretely} experience such services to be able to evaluate them, i.e.\ the domain can be simulated.

Service providers are modelled with probabilities associated with different outcomes, which are listed in Table~\ref{tbl:provider_model}. For example, providers are associated with a constant value that indicates the maximum days they take to deliver a package. They are also associated with a variable representing the average number of days that it takes to deliver packages and the standard deviation. Therefore, to simulate the number of days taken we used randomisation with a normal distribution defined by these parameters.

\begin{table}
	\centering
	\footnotesize
	\caption{Provider Model and Terms.}
	\begin{tabular}{ p{2.0cm} p{5.6cm} p{1.4cm} p{1.3cm} }
		\toprule
		\textbf{Outcome} & \textbf{Domain Values} & \textbf{Outcome} & \textbf{Term} \\
		& & \textbf{Model} & \\ 
		\midrule
		Number of{\par}days to deliver & Integer $> 0$ & Normal{\par}distribution{\par}$(\mu , \sigma)$ & Timeliness \\ \cline{1-3}
		Maximum{\par}days to deliver & Integer $> 0$ & Constant & \\ \hline
		Price & Double $> 0$ & Constant & Price \\ \hline
		Parcel{\par}Condition & Perfect Conditions{\par}Damaged Package{\par}Damaged Product{\par}Lost & Probabilities & Quality of{\par}Service \\ \hline
		Customer{\par}Service & Easy to contact and problem solved{\par}Easy to contact but problem unresolved{\par}Difficult to contact but problem solved{\par}Difficult to contact and problem unresolved  & Probabilities & Customer{\par}Support \\ \hline
		- & - & - & Reliability \\ 
		\bottomrule
	\end{tabular}
	\label{tbl:provider_model}
\end{table}%

Participants evaluate providers with respect to each term presented in the rightmost column of Table~\ref{tbl:provider_model}. These terms are associated with the outcome that we believe that the participant would take into account to rate a term. Note that reliability is not associated with any outcome, since we assume that this is related to repeated experiences that the participant has with the same provider.

We modelled 10 providers,  each being associated with two sets of model parameters. We use the first set of parameters to collect the first half of the set of sample ratings, and the second set of parameters to collect the remaining samples. In this way, we simulate change in the providers' behaviour, and allow for the fact that the ratings provided can change over time.

\subsection{Participants and Preferences}\label{sec:participants}

Our study participants were selected using \emph{convenience sampling}. Graduate and undergraduate students of a Brazilian Computer Science program were invited to participate as volunteers. Data was collected in two separate time slots, and participants that participated within the same time slot were considered peers, in order to compute witness trust. 
In total, our study involved \emph{30 participants}, such that 9 participated in the first time slot and 21 participated in the second. We detail characteristics of the participants in Table~\ref{tbl:participantCharacteristics}.

\begin{table}
	\caption{Characteristics of Participants (N = 30).}
	\centering
	\begin{tabular}{ l c c }
		\toprule
		\textbf{Age}	& 16--25 years	& 26--35 years	\\
						& 23 (77\%)		& 7 (23\%)		\\ \midrule
		\textbf{Gender}	& Male			& Female		\\
						& 29 (97\%)		& 1 (3\%)		\\ \midrule
		\textbf{Course}	& Undergraduate	& Graduate		\\
		\textbf{Level}	& 23 (77\%)		& 7 (23\%)		\\ 
		\bottomrule
	\end{tabular}
	\label{tbl:participantCharacteristics}
\end{table}%

In addition to collecting participant characteristics, we also asked them to provide their preferences with respect to reputation types and terms. Descriptive information was provided to allow them to understand the required information. In Table~\ref{tbl:participantPreferences}, we present the preferences provided by participants. Note that in this study we consider only interaction and witness reputation types.

\begin{table}
	\caption{Participant Preferences for Reputation Types and Terms.}
	\centering
	\begin{tabular}{ l r r }
		\toprule
		\textbf{Reputation Type/Term}	& \textbf{M}	& \textbf{SD}	\\ 
		\midrule
		Interaction						& 0.635			& 0.11 \\ 
		Witness							& 0.365			& 0.11 \\ 
		\midrule
		Customer Support				& 0.138			& 0.08 \\ 
		Price							& 0.202			& 0.09 \\ 
		Quality of Service				& 0.237			& 0.05 \\ 
		Reliability						& 0.242			& 0.07 \\ 
		Timeliness						& 0.181			& 0.07 \\ 
		\bottomrule	
	\end{tabular}
	\label{tbl:participantPreferences}
\end{table}%

\subsection{Results and Analysis}\label{sec:result}

We now present our study results, analysing first objective effectiveness and then perceived effectiveness. Hereafter explanation arguments and trust scores are referred to as \emph{arguments} and \emph{scores}, respectively.

\subsubsection{Objective Effectiveness}

The metric used to analyse objective effectiveness is the score difference between that given to explanatory and full information. Our aim is to evaluate collected scores in a single group but, because we had two separate participant groups (in order to obtain witness ratings), we first investigated whether results obtained are similar for both groups. We ran a Mann-Whitney's U test to compare group responses and, as expected, their is no significant difference between the scores provided by the two groups (U = 9436, p-value = 0.98).

\begin{figure}
	\centering
	\subfloat[Overall Score Differences.]{
		\includegraphics[width=0.75\linewidth]{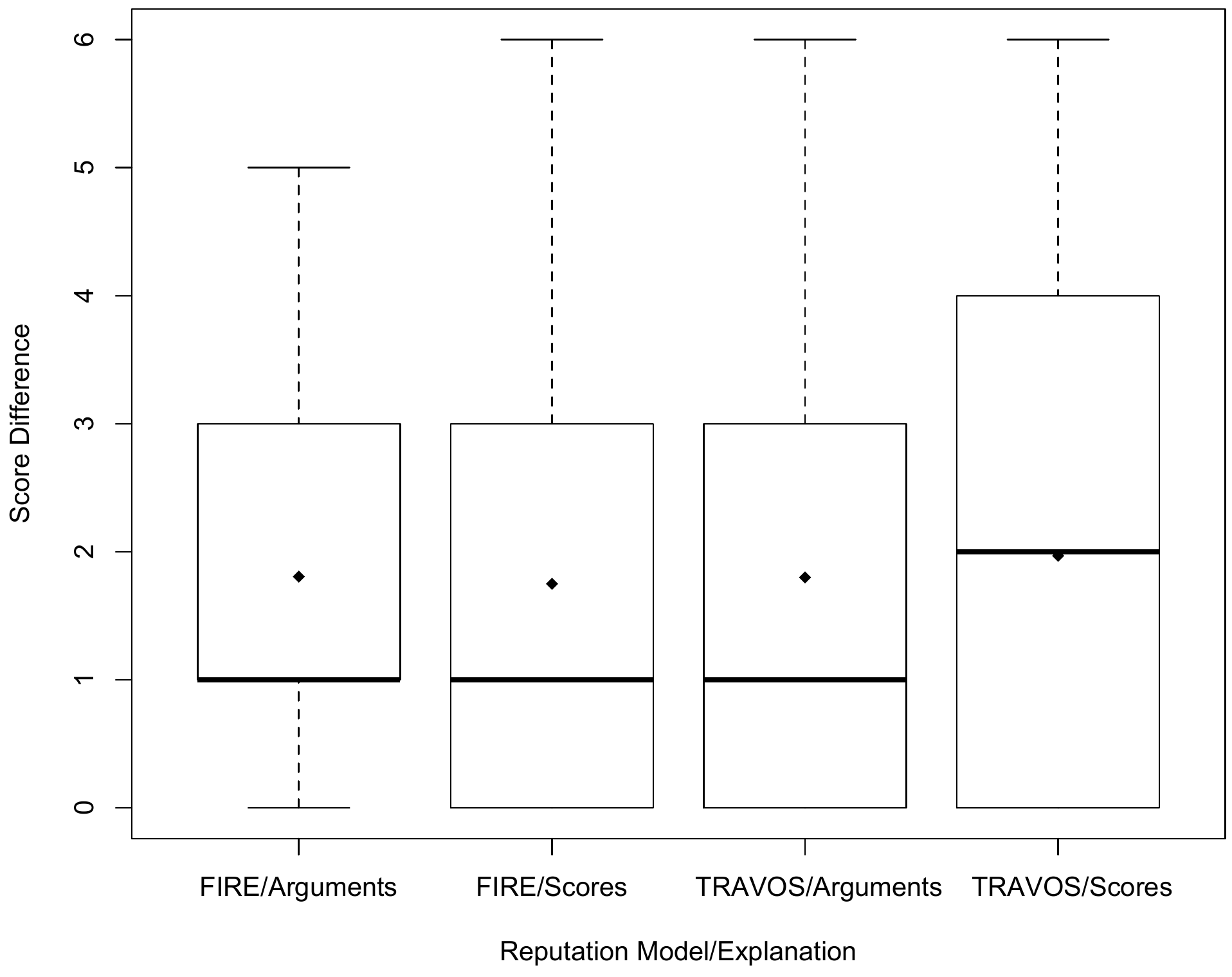}
			\label{fig:scoresOverall}}\\
	\subfloat[Score Differences by Agreement with the Model.]{\includegraphics[width=\linewidth]{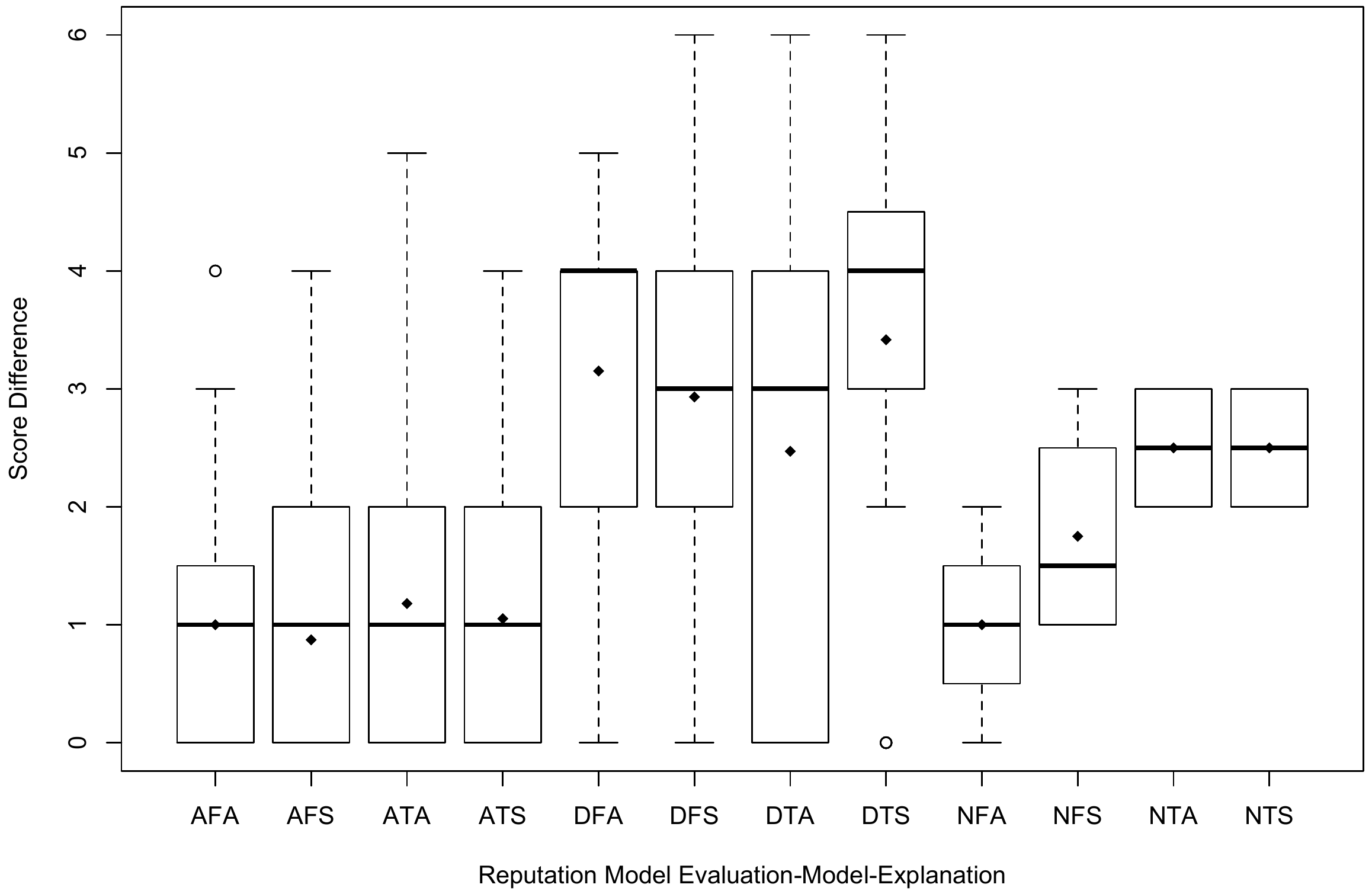}
		\label{fig:scoresByAgree}}
	\caption{Overview of Scores Differences.}
	\label{fig:explanationEvalScores}
\end{figure}

Considering participant scores, we obtained the results presented in the second (mean, M) and third (standard deviation, SD) columns of Table~\ref{tbl:scoreDifferences}.  Results are split into four groups (rows), according to the reputation model used (FIRE or TRAVOS) and the provided explanatory information (arguments or scores). Score differences for the four groups are also shown in Figure~\ref{fig:scoresOverall}, in a box plot, which presents the mean, median and variance of values. As can be seen, results diverge between FIRE and TRAVOS: while scores performed better considering FIRE, arguments outperformed scores considering TRAVOS. Despite these differences, a Kruskal-Wallis test revealed that the differences are not significant ($\chi^2$ = 13.7, p = 0.94). Although arguments and scores achieved similar results, this is already evidence of the effectiveness of our arguments. Arguments refer to a small portion of the information revealed by scores (it selects only decisive criteria, and provides further information only with respect to them). Therefore, we state our first finding as follows. 

\begin{framed}
\textbf{Finding 1:} Information that is not present in arguments can indeed be discarded, because it is not helpful to better evaluate providers, as otherwise using scores would have had a better performance. 
\end{framed}

Note that scores and arguments were presented separately in our study in order to understand the effectiveness of arguments in isolation, but we are not suggesting that this should be the case in real applications. We assume that they can be presented together, so that they can complement each other.

\begin{sidewaystable}
	\caption{Summary of Score Differences.}
	\centering
	\begin{tabular}{ l|c c|c c c c|c c c c c c }
		\toprule
		\textbf{Reputation Model/} & \multicolumn{2}{c|}{\textbf{Overall}}	& \multicolumn{4}{c|}{\textbf{Model-specific Arguments}}	& \multicolumn{6}{c}{\textbf{Agreement with the Model}} \\  	
		\textbf{Explanatory}	& \multicolumn{2}{c|}{}			& \multicolumn{2}{c}{\textbf{With}}	& \multicolumn{2}{c|}{\textbf{Without}} & \multicolumn{2}{c}{\textbf{Agree}}	& \multicolumn{2}{c}{\textbf{Disagree}}	& \multicolumn{2}{c}{\textbf{Neutral}} \\
		\textbf{Information}	& \textbf{M}	& \textbf{SD}	& \textbf{M}	& \textbf{SD}	& \textbf{M}	& \textbf{SD}	& \textbf{M}	& \textbf{SD}	& \textbf{M}	& \textbf{SD}	& \textbf{M}	& \textbf{SD}	\\ 
		\midrule
		FIRE/Arguments						& 1.81			& 1.53			& 1.76			& 1.52			& 1.88			& 1.57			 & 1.00			& 1.00			& 3.15			& 1.33			 & 1.00			& 0.82 \\ 
		FIRE/Scores							& 1.75			& 1.58			& -				& -				& -				& -				 & 0.87			& 1.03			& 2.93			& 1.51			 & 1.75			& 0.96 \\ 
		TRAVOS/Arguments					& 1.80			& 1.67			& 1.83			& 1.82			& 1.75			& 1.40			 & 1.18			& 1.10			& 2.47			& 1.97			 & 2.50			& 0.71 \\ 
		TRAVOS/Scores						& 1.97			& 1.79			& -				& -				& -				& -				 & 1.05			& 1.07			& 3.42			& 1.82			 & 2.50			& 0.71 \\ 
		\bottomrule
	\end{tabular}
	\label{tbl:scoreDifferences}
\end{sidewaystable}%

This initial analysis of our results showed that the differences among the four groups are not statistically significant. However, a deeper analysis allowed us to reveal interesting findings, which explain the contradicting results between FIRE and TRAVOS. First, we analysed whether the difference between the values obtained for FIRE and TRAVOS was due to the model quality, i.e., one model produces rankings that better match users’ opinions. Model quality is evaluated by checking whether the ranking produced by the reputation model matches the ranking that the users would produce, when they are aware of the full provider information. Consequently, in order to evaluate model quality, we used only the scores given by participants considering the full provider information. As shown in Table 6, rankings using trust scores calculated by FIRE and TRAVOS received similar ratings. Moreover, roughly, the same amount of participants agreed with the rankings produced by models. Indeed, Mann-Whitney’s U test indicates that the difference between the scores obtained with full information is not significant (U = 11482, p-value = 0.7).

Second, we investigated whether the model-specific arguments played a key role in our results. However, this was also not the case. In our results 61.96\% of the provided explanations contained model-specific arguments (61.36\% for FIRE, and 62.67\% for TRAVOS). In columns 6--7 of Table~\ref{tbl:scoreDifferences}, we detail the score differences between explanations provided with and without model-specific argument. We ran a Kruskal-Wallis test that showed that the differences are not significant ($\chi^2$ = 0.22, p = 0.97).

\begin{table}
	\caption{Quality of Reputation Assessment Models.}
	\centering
	\begin{tabular}{ l r r r r r}
		\toprule
		\textbf{Reputation Model}	& \textbf{M}	& \textbf{SD}	& \textbf{Agree}	& \textbf{Disagree} & \textbf{Neutral}	\\ 
		\midrule
		FIRE						& 4.48			& 2.00			& 56.25\%			& 38.75\%			& 5.00\%			\\ 
		TRAVOS						& 4.37			& 2.17 			& 55.71\%			& 41.43\%			& 2.86\%			\\ 
		\bottomrule
	\end{tabular}
	\label{tbl:modelQuality}
\end{table}%

We then analysed whether the agreement with model influenced the results. Scores were split into three groups: (i) \emph{agree}: when participants provided a score greater than 4 considering the full provider information, (ii) \emph{disagree}: when participants provided a score lower than 4, and (iii) \emph{neutral}: when participants provided a score equals to 4. Results are detailed in the last six columns of Table~\ref{tbl:scoreDifferences}. They are also shown in Figure~\ref{fig:scoresByAgree}, where the x-axis has labels with three letters: the first stands for \textbf{A}gree, \textbf{D}isagree, or \textbf{N}eutral, the second stands for \textbf{F}IRE or \textbf{T}RAVOS, and the third stands for \textbf{A}rguments or \textbf{S}cores. We observed that participants, in general, tend to agree with the ranking based on explanatory information, because this is only the information they have, which is in accordance with the ranking (the ranking is derived from scores).
Consequently, changes occur more often from agree to disagree than from disagree to agree, i.e., participants more often agree with the ranking considering explanatory information, and then change their opinion to disagree when they learn the full provider information.
A Kruskal Wallis test revealed a significant difference among the groups ($\chi^2$ = 97.7, p $<$ 0.01). A post-hoc test using Mann-Whitney tests with Bonferroni correction showed significant differences between the agree groups AFA, AFS and all disagree groups, and the agree groups ATA, ATS and the disagree groups (DFA, DFA, DTS). There is no significant difference between ATA and ATS, and DTA. This supports our second main finding.

\begin{framed}
\textbf{Finding 2:} Except arguments provided for TRAVOS (i.e.\ TRAVOS/Arguments), all combinations of reputation model with explanatory information (i.e.\ FIRE/Arguments, FIRE/Scores and TRAVOS/Scores) \emph{persuades} participants to agree with the ranking. 
\end{framed}

\begin{figure}
	\centering
	\includegraphics[width=\linewidth]{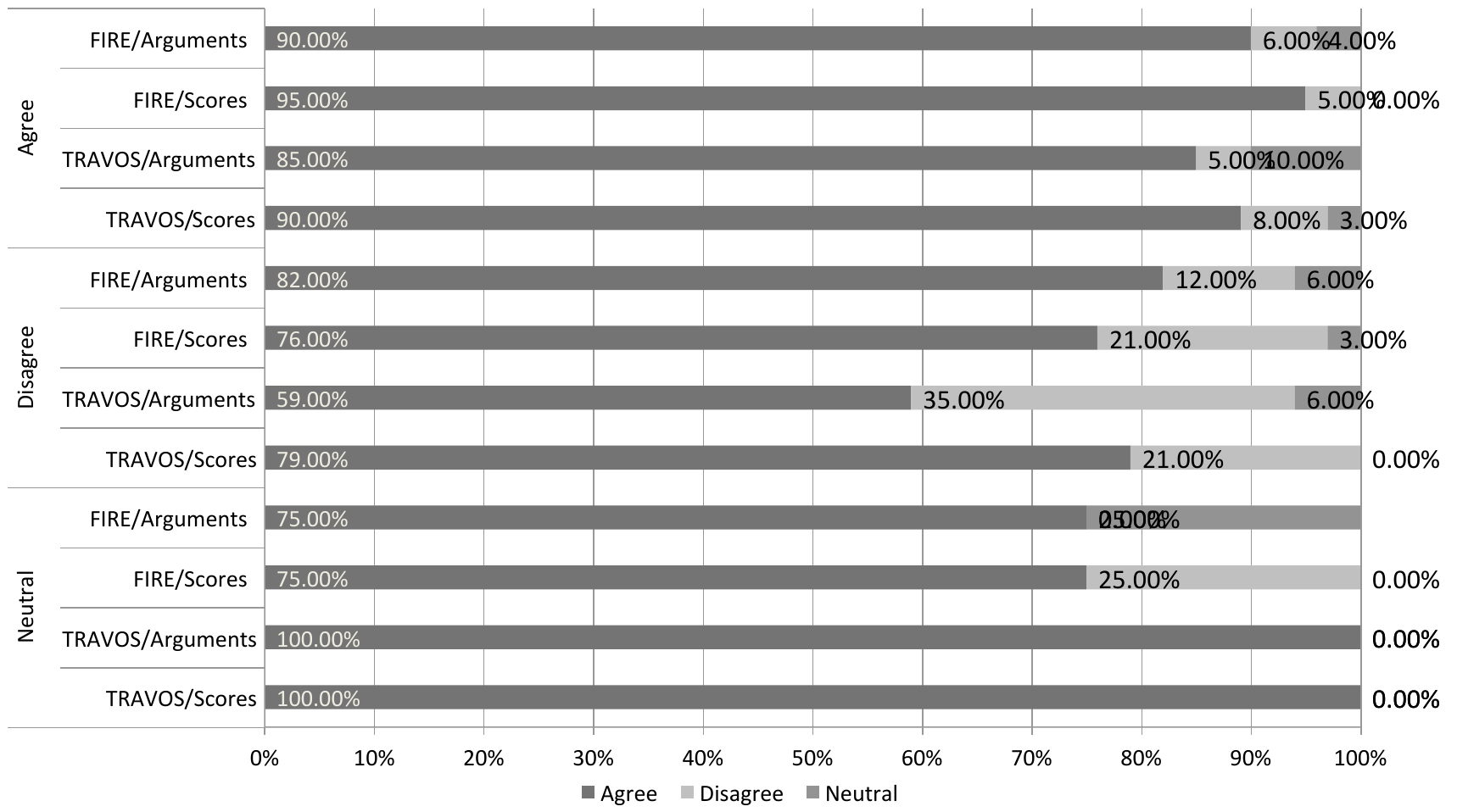}
	\caption{Distribution of evaluation based on explanatory information by agreement with the model.}
	\label{fig:opinionChangeDistribution}
\end{figure}

Our explanation approach thus managed to be not (or less) persuasive for one of the models, and this is a positive aspect of our approach. This result becomes evident in Figure~\ref{fig:opinionChangeDistribution}, in which we show the distribution of how participants evaluated the ranking based on explanatory information (divisions in columns shown in x-axis) according to how  they actually evaluate it, i.e.\ based on full provider information (y-axis). For example, from all cases in which participants evaluated FIRE/Arguments and they agreed with the model based on full provider information, in 90\% they agreed with the ranking based on explanatory information, in 6\% they disagreed with the model (when in fact they agree), and in 4\% they were neutral with the model. In most of the cases, participants agreed with the ranking based on explanatory information. Only with TRAVOS/Arguments, did they manage to more often perceive based on arguments that they actually disagree with the ranking (35\% of the cases).

We further investigated why this occurred, because this result is unexpected given that: (i) the reputation models are equally good, and (ii) explanation arguments are similar in FIRE and TRAVOS except for model-specific arguments, but explanations with model-specific arguments are not better than those without.
A key difference between FIRE and TRAVOS is that FIRE uses weights for reputation types that are given and TRAVOS calculates them, based on similarity between witnesses and interaction ratings. Consequently, even though the \emph{decisive reputation types} argument is used for both FIRE and TRAVOS, it reveals information of different nature. While in FIRE it just acknowledges participants that their preference for reputation types played a key role in the recommender, in TRAVOS it reveals a detail of the model that may be not in accordance with the participant preferences, e.g., the model gave importance to witnesses opinions while the participant believes that such opinions are not that important. Therefore, our hypothesis that explains this result leads to our third finding.

\begin{framed}
\textbf{Finding 3:} Arguments that reveal implicit model information, which is the result of a calculation or an assumption regarding user preferences, are essential for users to better understand the rationale behind reputation assessments and use such information to make better decisions.
\end{framed}

\subsubsection{Perceived Effectiveness}

\begin{figure}
	\centering
	\includegraphics[width=\linewidth]{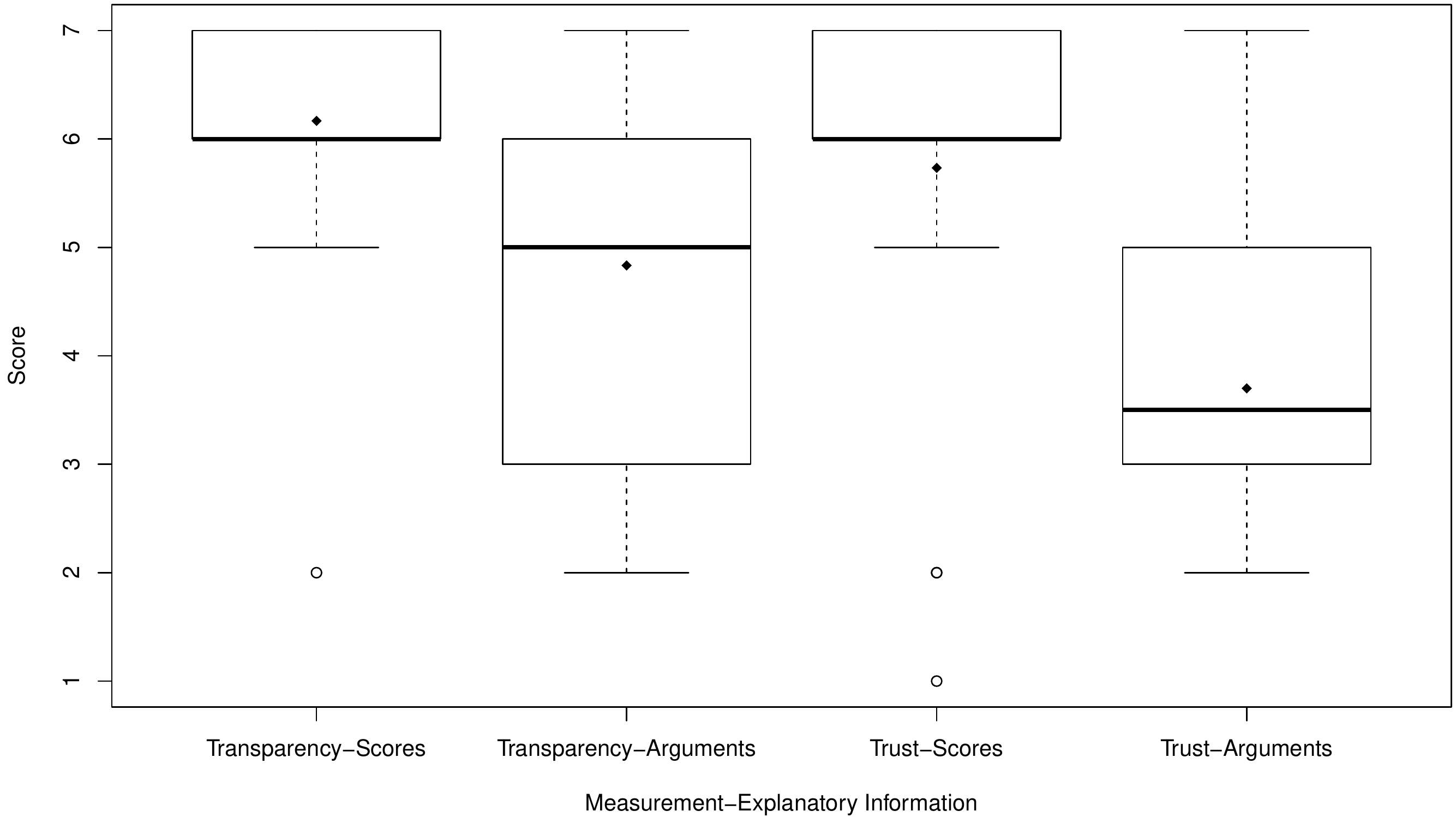}
	\caption{Questionnaire Scores: Transparency and Trust.}
	\label{fig:questionnaireScores}
\end{figure}

In addition to the evaluation of the objective effectiveness of our approach, we also analysed how participants perceive the explanations. Results with respect to transparency and trust in our explanations, presented in Figure~\ref{fig:questionnaireScores}, show that participants prefer scores instead of textual explanations. A Wilcoxon Signed-ranks test indicated that the difference between scores (M = 6.17; SD = 1.02) and arguments (M = 4.83; SD = 1.53) with respect to transparency is statistically different (W = 25.5; p $<$ 0.01), and the difference between scores (M = 5.73; SD = 1.53) and arguments (M = 3.70; SD = 1.39) with respect to transparency is also statistically different (W = 58.5; p $<$ 0.01). This result was expected given that our explanation arguments, when translated to a textual form, requires the user to read a possibly large set of sentences, and a previous study~\cite{Nunes:ECAI2014:ExplanationTechnique} showed that this may cause users to dislike it. Based on this, we state our fourth and last finding.

\begin{framed}
\textbf{Finding 4:} It is important to identify graphical forms of presenting the information captured by our explanation arguments.
\end{framed}

Note that although participants indicated that scores were more transparent than explanation arguments, as shown, they are similarly effective and arguments are less persuasive under certain circumstances. Moreover, even though lengthy explanations are criticised by participants, they do not impact on effectiveness or efficiency. This is shown in Figure~{fig:explanationLength}, which shows the lack of correlation between the explanation length and score differences (effectiveness) and time to analyse them (efficiency). The results of our subjective analysis, however, provide evidence of the need for better means of translating our explanation arguments into a human-readable presentation format.

\begin{figure}
	\centering
	\subfloat[Explanation Length vs.\ Score Difference.]{\includegraphics[width=\linewidth]{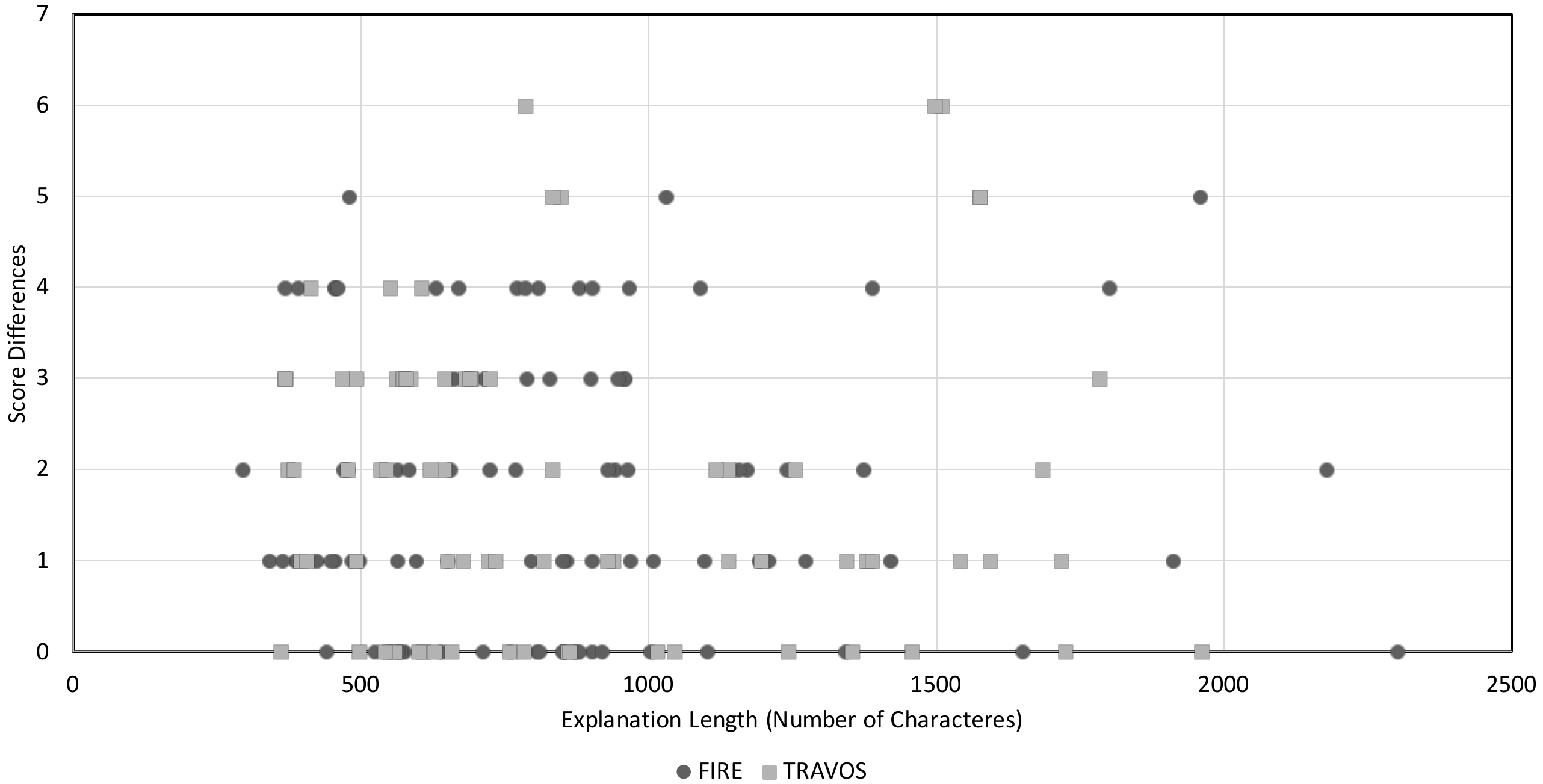}		
		\label{fig:explanationLengthTimeScores}}\\
	\subfloat[Explanation Length vs.\ Time.]{\includegraphics[width=\linewidth]{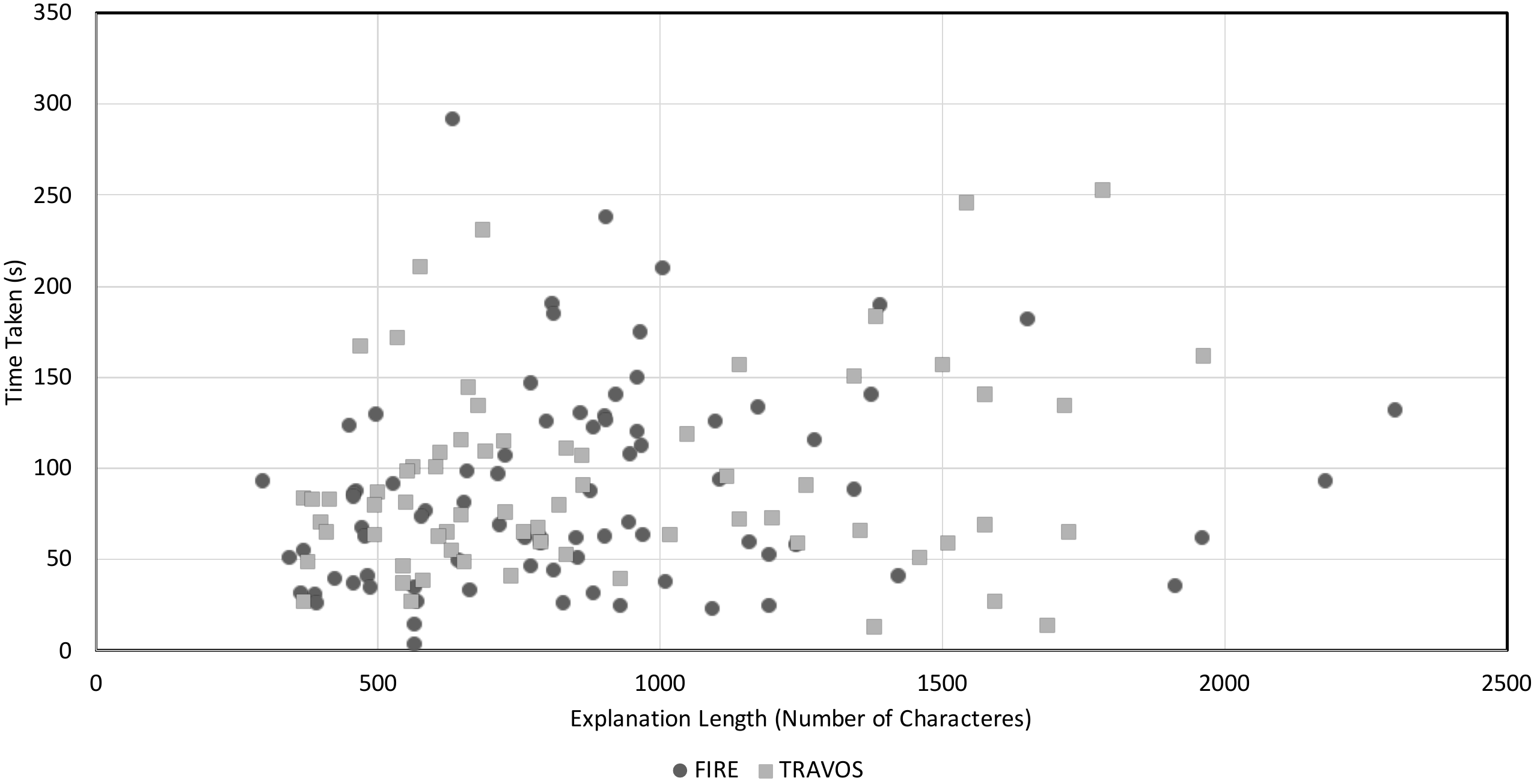}
		\label{fig:explanationLengthTime}}
	\caption{Analysis of the Impact of Explanation Length.}
	\label{fig:explanationLength}
\end{figure}

Interestingly, some participants did not realise that the textual-based explanations were explaining the scores, and believed that the arguments were trying to convince them to agree with the ranking. When justifying their transparency and trust scores, five participants reported that textual explanations can persuade them and scores cannot, mainly because they can see the exact difference between scores, but our results show that this is not the case. In fact, as discussed in the related work section, a study concluded that showing ratings from neighbours can persuade users to accept recommendations \cite{Herlocker:CSCW2000:ExplainingRecommender}, so this previous study and ours converge to the same direction. Four participants highlighted benefits of our arguments, such as providing meaning to small quantitative differences or analysing recency. One of the participants made the following comment:
\emph{``The explanations with scores can [be] ambiguous sometimes, specially when scores differ on small amounts e.g How much is 0.002 of reliability? However, textual explanations not only remove that ambiguity, but also make certain aspects of the ordering explicit, such as your personal weights, and recent scoring being more important than overall, for example.''} Finally, two participants reported that although they prefer scores, the textual explanations provide complementary information, which is the main aim in our case. 

%% file: sec6_conclusion.tex

\section{Conclusion}\label{sec:conclusion}

In this paper, we have presented an approach to generating explanations of why providers of services were considered to have more or less reputation than other providers. This involved abstracting existing reputation assessment models into a generalised model that we used as a base to produce explanations. In our work, we leveraged existing explanation approaches (for multi-attribute decision models) to determining decisive arguments when choosing between options, to account for the different values that are weighted in reputation assessment, such as the weighting between a client's own past experience and the information it has gathered from its peers. We presented a model by which concise arguments could be extracted from the reputation assessment process and combined into explanations. 
Explanation arguments were evaluated with a user study. We concluded that, although explanations present a subset of the information of trust scores, they are sufficient to equally evaluate providers recommended based on their trust score. Moreover, when explanation arguments reveal implicit model information, they are less persuasive than scores. Despite these positive aspects of our explanations, given that they are presented in a textual form, which requires more cognitive effort to analyse, participants showed preference for analysing scores instead of reading sentences.

For illustration, we have considered in this paper the FIRE and TRAVOS reputation models. However, our approach is unchanged if an alternative reputation model is adopted, as long as it can be mapped to our generalised multi-term reputation model. We do not assume a particular representation of behaviour or source of information, nor require a particular method of assessing reputation from available sources. We identify the overall decisive criteria for a provider being preferred to another, and subsequently identify the corresponding model-specific arguments that support the assessment. The process of identifying the criteria and generating explanations is unchanged, but the details of the criteria may be different, e.g.\ criteria for ReGreT~\cite{Sabater-Mir:2001aa,Sabater:2004aa} might consider trust ascribed to the groups to which agents belong, while for HABIT~\cite{Teacy:2012aa} the criteria would refer to probabilistic estimations of future behaviour.

We currently focused on using and evaluating our approach with human users. However, automated negotiation environments can also potentially benefit from our explanations. For example, when automated providers are selected (or not selected) by clients, they can ask for explanations to help them improve their services. In addition, explanations can be used by clients to improve their choices by refining their preferences. 
Clients may also use explanations to change their network neighbours. If a client observes that it always chooses providers because they are better rated considering its own experience, even though ratings given by peers are higher, the client may understand that its ratings diverge from its peers, and possibly look for new neighbours. Moreover, explanations may be used to share information among clients. For instance, a client concerned with privacy issues can state to other clients which provider is better than another using an explanation as a rationale, without revealing their preferences and ratings. All these different directions will be explored in our future work.

%% file: ijhcs-2017.bbl
\begin{thebibliography}{10}
\expandafter\ifx\csname url\endcsname\relax
  \def\url#1{\texttt{#1}}\fi
\expandafter\ifx\csname urlprefix\endcsname\relax\def\urlprefix{URL }\fi
\expandafter\ifx\csname href\endcsname\relax
  \def\href#1#2{#2} \def\path#1{#1}\fi

\bibitem{Ricci:2010:RSH:1941884}
F.~Ricci, L.~Rokach, B.~Shapira, P.~B. Kantor, Recommender Systems Handbook,
  1st Edition, Springer-Verlag New York, Inc., New York, NY, USA, 2010.

\bibitem{Ramchurn:2004aa}
S.~D. Ramchurn, T.~D. Huynh, N.~R. Jennings, Trust in multi-agent systems, The
  Knowledge Engineering Review 19~(1) (2004) 1--25.

\bibitem{Sabater:2004aa}
J.~Sabater, Evaluating the {ReGreT} system, Applied Artificial Intelligence
  18~(9-10) (2004) 797--813.

\bibitem{Huynh:2006aa}
T.~D. Huynh, N.~R. Jennings, N.~R. Shadbolt, An integrated trust and reputation
  model for open multi-agent systems, Journal of Autonomous Agents and
  Multi-Agent Systems 13~(2) (2006) 119--154.

\bibitem{Teacy:2005aa}
W.~T.~L. Teacy, J.~Patel, N.~R. Jennings, M.~Luck, Coping with inaccurate
  reputation sources: Experimental analysis of a probabilistic trust model, in:
  Proceedings of the 4th International Conference on Autonomous Agents and
  Multiagent Systems, 2005, pp. 997--1004.

\bibitem{Regan:2006aa}
K.~Regan, P.~Poupart, R.~Cohen, Bayesian reputation modeling in e-marketplaces
  sensitive to subjectivity, deception and change, in: Proceedings of the 21st
  National Conference on Artificial Intelligence, 2006.

\bibitem{Teacy:2012aa}
W.~T.~L. Teacy, M.~Luck, A.~Rogers, N.~R. Jennings, An efficient and versatile
  approach to trust and reputation using hierarchical bayesian modelling,
  Artificial Intelligence 193 (2012) 149--185.

\bibitem{Tintarev:chapter2011:Explanation}
N.~Tintarev, J.~Masthoff, Designing and evaluating explanations for recommender
  systems, in: Recommender Systems Handbook, Springer US, 2011, pp. 479--510.

\bibitem{Labreuche:AI2011:Explanations}
C.~Labreuche, A general framework for explaining the results of a
  multi-attribute preference model, Artif. Intell. 175~(7-8) (2011) 1410--1448.
\newblock \href {http://dx.doi.org/10.1016/j.artint.2010.11.008}
  {\path{doi:10.1016/j.artint.2010.11.008}}.

\bibitem{Nunes:ECAI2014:ExplanationTechnique}
I.~Nunes, S.~Miles, M.~Luck, S.~Barbosa, C.~Lucena, Pattern-based explanation
  for automated decisions., in: Proceedings of the 21th Eureopean Conference on
  Artificial Intelligence, ECAI'2014, 2014, pp. 669--674.

\bibitem{bilgic:iui05-wkshp}
M.~Bilgic, R.~Mooney,
  \href{http://www.cs.iit.edu/~ml/pdfs/bilgic-iui05-wkshp.pdf}{Explaining
  recommendations: Satisfaction vs. promotion}, in: Proceedings of Beyond
  Personalization 2005: A Workshop on the Next Stage of Recommender Systems
  Research at the 2005 International Conference on Intelligent User Interfaces,
  2005.
\newline\urlprefix\url{http://www.cs.iit.edu/~ml/pdfs/bilgic-iui05-wkshp.pdf}

\bibitem{Teacy:2006aa}
W.~T.~L. Teacy, J.~Patel, N.~R. Jennings, M.~Luck, {TRAVOS}: {T}rust and
  reputation in the context of inaccurate information sources, Journal of
  Autonomous Agents and Multi-Agent Systems 12 (2006) 183--198.

\bibitem{Herlocker:CSCW2000:ExplainingRecommender}
J.~L. Herlocker, J.~A. Konstan, J.~Riedl, Explaining collaborative filtering
  recommendations, in: CSCW '00, ACM, New York, NY, USA, 2000, pp. 241--250.

\bibitem{Carenini:AI2006:EvaluativeArguments}
G.~Carenini, J.~D. Moore, Generating and evaluating evaluative arguments,
  Artif. Intell. 170 (2006) 925--952.

\bibitem{Ye:MISQ1995:ExplanationImpact}
L.~R. Ye, P.~E. Johnson, The impact of explanation facilities on user
  acceptance of expert systems advice, MIS Q. 19 (1995) 157--172.
\newblock \href {http://dx.doi.org/http://dx.doi.org/10.2307/249686}
  {\path{doi:http://dx.doi.org/10.2307/249686}}.

\bibitem{Gedikli:2014:IEC:2580118.2580448}
F.~Gedikli, D.~Jannach, M.~Ge, How should {I} explain? a comparison of
  different explanation types for recommender systems, Int. J. Hum.-Comput.
  Stud. 72~(4) (2014) 367--382.
\newblock \href {http://dx.doi.org/10.1016/j.ijhcs.2013.12.007}
  {\path{doi:10.1016/j.ijhcs.2013.12.007}}.

\bibitem{Tintarev:RecSys2007:EffectiveExplanations}
N.~Tintarev, J.~Masthoff, Effective explanations of recommendations:
  user-centered design, in: Proc. of the 2007 ACM conference on Recommender
  systems, RecSys '07, USA, 2007, pp. 153--156.

\bibitem{Nunes:2012:IEJ:2358968.2358987}
I.~Nunes, S.~Miles, M.~Luck, C.~J.~P. de~Lucena, Investigating explanations to
  justify choice, in: Proceedings of the 20th International Conference on User
  Modeling, Adaptation, and Personalization, UMAP'12, Springer-Verlag, Berlin,
  Heidelberg, 2012, pp. 212--224.

\bibitem{Carenini:IJCAI2001:ExplanationStudy}
G.~Carenini, J.~D. Moore, An empirical study of the influence of user tailoring
  on evaluative argument effectiveness, in: Proceedings of the 17th
  international joint conference on Artificial intelligence, IJCAI'01, USA,
  2001, pp. 1307--1312.

\bibitem{Klein:AA1994:ExplanationFW}
D.~A. Klein, E.~H. Shortliffe, A framework for explaining decision-theoretic
  advice, Artif. Intell. 67 (1994) 201--243.
\newblock \href {http://dx.doi.org/10.1016/0004-3702(94)90053-1}
  {\path{doi:10.1016/0004-3702(94)90053-1}}.

\bibitem{Briguez12}
C.~Briguez, M.~Bud{\'a}n, C.~Deagustini, A.~Maguitman, M.~Capobianco,
  G.~Simari, Towards an argument-based music recommender system, in: Frontiers
  in Artificial Intelligence and Applications, Vol. 245, 2012, pp. 83--90.

\bibitem{RecioGarcia13}
J.~A. Recio-Garc\'{\i}a, L.~Quijano, B.~D\'{\i}az-Agudo, Including social
  factors in an argumentative model for group decision support systems,
  Decision Support Systems 56 (2013) 48--55.

\bibitem{BRIGUEZ14}
C.~E. Briguez, M.~C. Bud{\'a}n, C.~A. Deagustini, A.~G. Maguitman,
  M.~Capobianco, G.~R. Simari, Argument-based mixed recommenders and their
  application to movie suggestion, Expert Systems with Applications 41~(14)
  (2014) 6467 -- 6482.

\bibitem{Chesnevar09}
C.~Ches{\~{n}}evar, A.~G. Maguitman, M.~P. Gonz{\'a}lez, Empowering
  recommendation technologies through argumentation, in: G.~Simari, I.~Rahwan
  (Eds.), Argumentation in Artificial Intelligence, Springer US, Boston, MA,
  2009, pp. 403--422.

\bibitem{Rodriguez16}
P.~Rodr{\'i}guez, S.~Heras, J.~Palanca, N.~Duque, V.~Juli{\'a}n,
  Argumentation-based hybrid recommender system for recommending learning
  objects, in: M.~Rovatsos, G.~Vouros, V.~Julian (Eds.), Multi-Agent Systems
  and Agreement Technologies, Springer International Publishing, Cham, 2016,
  pp. 234--248.

\bibitem{Garcia04}
A.~J. Garc\'{\i}a, G.~R. Simari, Defeasible logic programming: An argumentative
  approach, Theory and Practice of Logic Programming 4~(2) (2004) 95--138.

\bibitem{Sabater:2005aa}
J.~Sabater, C.~Sierra, Review on computational trust and reputation models,
  Artificial Intelligence Review 24~(1) (2005) 33--60.

\bibitem{Gambetta:1988aa}
D.~Gambetta, Can we trust trust?, in: D.~Gambetta (Ed.), Trust: Making and
  Breaking Cooperative Relations, Oxford: Basil Blackwell, 1988, pp. 213--237.

\bibitem{Josang:2007aa}
A.~J{\o}sang, R.~Ismail, C.~Boyd, A survey of trust and reputation systems for
  online service provision, Decision Support Systems 43 (2007) 618--644.

\bibitem{Pinyol:2013aa}
I.~Pinyol, J.~Sabater-Mir, Computational trust and reputation models for open
  multi-agent systems: a review, Artificial Intelligence Review 40 (2013)
  1--25.

\bibitem{Wang:2007aa}
Y.~Wang, M.~P. Singh, Formal trust model for multiagent systems, in:
  Proceedings of the 20th International Joint Conference on Artificial
  Intelligence, 2007, pp. 1551--1556.

\bibitem{Sabater-Mir:2001aa}
J.~Sabater-Mir, C.~Sierra, Regret: A reputation model in gregarious societies,
  in: Proceedings of the 4th Workshop on Deception, Fraud and Trust in Agent
  Societies, 2001, pp. 61--69.

\bibitem{Burnett:2013aa}
C.~Burnett, N.~Oren, Position-based trust update in delegation chains, in:
  Proceedings of the 16th International Workshop on Trust in Agent Societies,
  2013.

\bibitem{Sensoy:2016aa}
M.~{\c{S}}ensoy, B.~Yilmaz, T.~J. Norman, {STAGE}: Stereotypical trust
  assessment through graph extraction, Computational Intelligence 32~(1) (2016)
  72--101.

\bibitem{Whitby:2004aa}
A.~Whitby, A.~J{\o}sang, J.~Indulska, Filtering out unfair ratings in
  {B}ayesian reputation systems, in: Proceedings of the Workshop on Trust in
  Agent Societies at AAMAS 2004, 2004.

\bibitem{Josang:2002aa}
A.~J{\o}sang, R.~Ismail, The beta reputation system, in: Proceedings of the
  15th Bled Electronic Commerce Conference e-Reality: Constructing the
  e-Economy, 2002, pp. 324--337.

\bibitem{Keeney:book1976:MAUT}
R.~L. Keeney, H.~Raiffa, Decisions with Multiple Objectives: Preferences and
  Value Tradeoffs, Wiley series in probability and mathematical statistics,
  John Wiley \& Sons, Inc, New York, 1976.

\end{thebibliography}
